\documentclass{article}
\usepackage[dvips]{graphicx}
\usepackage{subfigure}
\usepackage{caption}

\usepackage{amsmath}
\usepackage{amssymb}
\usepackage{array}
\usepackage{url}

\usepackage{xcolor}

\newcommand{\MLEn}{\hat{\theta}_{n}}
\newcommand{\MLE}{\hat{\theta}}
\newcommand{\Like}{\ensuremath{\mathcal{L}}}

\newcommand{\logL}{\ensuremath{\ln\mathcal{L}}}
\DeclareMathOperator\sign{sign}

\usepackage[auth-sc-lg]{authblk}

\begin{document}

\title{Fundamental Issues Regarding Uncertainties in Artificial Neural Networks.}

\author[$\star$]{N. A. Thacker}
\author[$\diamond,\star$]{C. J. Twining}
\author[$\star$]{P. D. Tar}
\author[$\dag$]{S. Notley}
\author[$\ddag$]{V. Ramesh}
\affil[$\star$]{Division of Informatics, Imaging \& Data Sciences, University of Manchester, U.K.}
\affil[$\diamond$]{Department of Computer Science, University of Manchester, U.K.}
\affil[$\dag$]{Department of Automatic Control and Systems Engineering, University of Sheffield, U.K.}
\affil[$\ddag$]{Center for Cognition and Computation, Goethe-Universit\"{a}t, Frankfurt am Main, Germany}
\date{}

\maketitle

\begin{abstract}\textsl{
Artificial Neural Networks (ANNs) implement a specific form of multi-variate
extrapolation and will generate an output for any input pattern, even when there is
no similar training pattern.
Extrapolations are not necessarily to be trusted, and in order to support safety critical systems, we require  such systems to give an
indication of the training sample related uncertainty associated
with their output. Some readers may think that this is a well known issue
which is already covered by the basic principles of pattern recognition.
We will explain below how this is not the case and how the conventional
(Likelihood estimate of) conditional probability of classification
does not correctly assess this uncertainty.
We provide a discussion of the standard interpretations of
this problem and show how a quantitative approach based upon long standing
methods can be practically applied.
The methods are illustrated on the task of early diagnosis of dementing diseases
using Magnetic Resonance Imaging.}
\end{abstract}

\section{Introduction}

Machine learning, and in particular artificial neural networks, have been applied successfully in a number of areas with state-of-the-art performance \cite{Alom2019}. A key research challenge, identified by The Royal Society \cite{RoySoc2017}, is verification and robustness, especially for safety-critical applications, where the quality of decisions and predictions must be verifiable to a high standard.
This high standard of robustness must be maintained, not only in the large scale/big data scenario, but also in applications where only smaller amounts of labelled data are available.

Conventional descriptions of pattern recognition systems
relate the output of ANN's to conditional
probabilities of classification $P(C|{\bf X})$ ~\cite{Lippmanb}.
Although it would be convenient to assume that
this output tells us something useful about uncertainty,
in reality it does not. The problem here arises from
the density of samples (${\bf X}$) in the vicinity of the input pattern.
When there is only one pattern from which to determine the
output, $P(C|{\bf X})$ will be driven to a value of 0 or 1.  Whereas
from a statistical perspective the sample size is simply not large
enough to be so confident. In order to really
understand our output we need to know not only $P(C|{\bf X})$
but also the total sample density which gave rise to it. This makes
it possible to check that output data will support decision making at a level
which meets performance specifications ~\cite{Haralick}.
One way to explain this is to say that
the system output is the maximum Likelihood estimate of $P(C|{\bf X})$,
whereas what we need to know is the expectation value $E\left[P(C|{\bf X})\right]$.
We will illustrate this difference below for Binomial statistics,
which is the simplest sample based probability estimate.

Some have tackled
this problem using a ``what if'' approach, where an effort is made
to identify the specific pieces of information which have most influence ~\cite{Horel}.
Bishop ~\cite{Bishop1993} considered the problem of validating outputs from a multi-layer perceptron by explicitly modelling the density of the input space using a Parzen window \cite{Parzen1962} based approach. For, areas of the input space with low density the outputs are flagged as unreliable. In a similar approach, based on radial basis function networks, Leonard \emph{et al} \cite{Leonard1992a,Leonard1992b} use the hidden nodes of the network as the model of the input space density.
As well as flagging unreliable outputs due to low input density, the method attempt to put 95\% confidence intervals on the outputs. This is based on Student’s t-statistics, for 95\% confidence, of the cross-validation error, with the number of degree of freedom given by the number of input vectors that significantly activate contributing hidden nodes.

Uncertainties arising from artificial decision systems can be grouped into three categories, statistical uncertainties (due to perturbations in input data),
systematic uncertainties (due to the uncertainties associated with training) and bias
(due to use of a mis-specified functional model).
Ideally to understand the reliability of an output we need all of these.
Kendall and Gal \cite{Kendall2017} refer to processes
of aleotoric and epistemic uncertainty, of which the former is the statistical error and
the latter is at least the systematic error. In this paper will assume that epistemic error
is systematic error and discuss bias as a separate issue.

The statistical uncertainty can be obtained in a relatively straightforward manner, by perturbing the input and observing the consequent variation in output. Numerical approaches based upon error prorogation can even make use of the derivatives used during training to make such assessments.
In the earlier work of Gal \cite{Gal2016a} statistical uncertainty is modelled via a subjective belief in the smoothness of the output function. This prior assumption has a constraining effect similar to the covariance function used in Gaussian processes. In the more recent work \cite{Kendall2017}, this term has been replaced as the amount of noise inherent in the observed output data, and is either tuned (for homoscedastic noise) or learned as function of the input data for heteroscedastic noise.

In order to estimate the systematic (epistemic) uncertainty associated with a predictor we need first either a quantitative description of the possible variations in training data, or a description of the uncertainty on the associated parameters (consistent with the former). Variations over data may be computationally intensive to assess. If it is possible to obtain the uncertainties in parameters then any
subsequent estimation of output uncertainty is likely to be more efficient, as the number of trained parameters should always be less than the number of training patterns.

A recent approach to this has been suggested by Gal and Ghahramani \cite{Gal2016a, Gal2016b}, using drop-out \cite{Srivastava2014} in a monte carlo style approach, to estimate uncertainty on the outputs. It is shown that training a neural network with drop out is mathematically equivalent to a Gaussian process \cite{Rasmussen2006}. Drop out is used as computationally efficient approximation to variational inference \cite{Graves2011, Blundell2015, Hinton1993} without increasing the number of model parameters. However, the equivalence only holds for large numbers of hidden nodes and is thus not fully scalable or universally valid.

Predictive uncertainty, due to systematic uncertainty on the weights is obtained, using drop-out as a Monte Carlo integration, by estimating the first two raw moments, under the assumption that the joint density of the outputs are diagonal multivariate normal. In this case, the number of terms in the integration is limited by the number of weights, and only appropriate for large scale networks. Even with large scale networks, the multivariate joint density of the weights is not fully sampled and correlations between parameters, which may be significant \cite{Barber1998}, are not accounted for. In more recent work \cite{Gal2017}, the discrete form of drop-out is relaxed to a continuous Concrete distribution \cite{Maddison2017} with improved uncertainty estimates when compared to synthetic data with known uncertainties.

We wish to make explicit the amount of epistemic uncertainty associated with any decision system
and the consequent uncertainty then arising during use.
We will tackle the problem using approaches more closely related to
formal statistics than the work cited above (see for example ~\cite{Murata}), but will explain below
how earlier methods were restricted both by numerical practicalities and unrealistic approximations.
In our opinion, a quantitative statistical approach for the assessment of epistemic uncertainty
would involve sampling of weights from their expected uncertainty distribution
whilst maintaining a high standard of validation, even for cases involving small data and/or networks.

The correct interpretation of training cost functions is Likelihood,
and its maximisation is the de-facto method for parameter estimation.
However, even when taking steps to ensure that the Likelihood
construct is a valid statistical description of data uncertainty (i.e. honest ~\cite{Dawid}),
the variation of the Likelihood function over the parameter has the wrong
properties to allow it to be interpreted directly as a parameter density.
Regardless of the choice of probability framework (Bayesian or
Frequentist) it is accepted that a Likelihood function needs to be multiplied
by something which is a function of the parameters in order to construct
a consistent (Bayesian) or quantitatively valid (Frequentist) description
of uncertainty. Therefore, our first task is to identify the principles
which specify how this should be done.

\subsection{Background Theory \& Notation}

We introduce here the theory associated with estimation of parameter
uncertainty when using Likelihood in order to identify the
relationships between competing methodologies and associated approximations.
Suppose we have a dataset of $n$ i.i.d. observations: $\mathbf{X} = \{X_{i}:i=1,\ldots n\} = \{X_{1},\ldots,X_{n}\}$, with an associated family of parametric
conditional pdf(s)
$p(X_{i}|\theta)$. We hence can write the pdf for the entire dataset as:
\mbox{$p(\mathbf{X}|\theta) \equiv \prod\limits_{i=1}^{n}p(X_{i}|\theta)$}. The Likelihood and log-likelihood for the observations are then:
\begin{equation}
\Like(\theta,\mathbf{X}) \equiv \Like_{n}(\theta) =
p(\mathbf{X}|\theta), \:\: l_{n} = \logL(\theta,\mathbf{X}), \:\: \mbox{and we define} \:\:\: \hat{\theta}(\mathbf{X})\:\: \mbox{s.t.} \:\: \left.\frac{\partial \Like(\theta,\mathbf{X})}{\partial \theta}\right|_{\theta=\hat{\theta}} = 0,
\end{equation}
where $\hat{\theta}(\mathbf{X})$ or $\MLEn$ is the maximum-likelihood estimator for the parameter. The standard Bayesian approach is to construct a posterior distribution over the space of parameters conditioned on the data by means of a prior distribution on parameter space $\pi(\theta)$ thus (see Jeffreys~\cite{jeffreysbook}, Theorem 10):
\begin{equation}
\pi^*(\theta|\mathbf{X}) \propto \Like(\theta,\mathbf{X})\pi(\theta). \label{eqn:postdef}
\end{equation}
Ideally, in the absence of any other meaningful estimate of the priors,
we need an uninformative prior. Unfortunately, {\bf this is not as simple as
making $\pi({\bf \theta})$ uniform} (e.g. as implicitly assumed
in ~\cite{Murata,Anders}), as this makes a fundamental (and
unfortunately common) error regarding the correct use of probabilities
and probability densities{\footnote{An uninformative prior
{\it probability} is flat, an uninformative {\it density} is not, and for
arbitrary parameters, see below, $\pi$ is the latter. This difference is hard to comprehend if no distinction is made between the two.}}.
Alternatively, a Jeffreys prior can also be constructed by using his `general rule', and requiring that the functional form of the prior is invariant under arbitrary transformations of the parameter space~\cite{jeffreys1946invariant}.

However, we here instead give a na\"{\i}ve derivation for the case of a single parameter, in order to make clear the link between the Jeffreys priors under the Bayesian approach, and the approach taken by Welch and by Peers~\cite{Welch1963,Peers1965}, which leads to approaches that seek to achieve a Bayesian-Frequentist synthesis~\cite{Ghosh2011,Kass}.
We will then exploit this interpretation to develop
a practical method for use with highly non-linear systems, i.e. ANNs.

\subsubsection{Interpretation of Jeffreys Priors.}
We show below how the Jeffreys prior is
related to the process of re-mapping parameters to achieve Gaussian
Likelihood functions.
We start with the 1-parameter log-likelihood $l_{n}(\theta)$ or Likelihood $\Like_{n}(\theta)$ for $n$ data points, and hence the posterior:
\[
\pi^{*}_{n}(\theta) \propto \Like_{n}(\theta)\pi(\theta).
\]
We will consider the case where $n$ is large, and we will assume that the Likelihood is then strongly-peaked about the maximum likelihood estimate (MLE) of the parameter $\MLEn$. We then first shift and scale to define a new parameter, $\theta \rightarrow t$, where:
\begin{equation}
t \doteq \sqrt{n}(\theta-\MLEn)\hat{I}_{n}^{\frac{1}{2}}, \:\: \mbox{and:} \:\: \hat{I}_{n} = -\frac{1}{n}\left.\frac{d^{2}l_{n}}{d\theta^{2}}\right|_{\theta=\MLEn}.\label{eqn:FINdef}
\end{equation}
We here have introduced the empirical Fisher Information function per data point $I_{n}$, which has been replaced  by its value at the optimum $\hat{I}_{n} \equiv I_{n}(\MLEn)$. We see that the new variable is centred on the peak of the Likelihood, and the width of the peak has been scaled by the second derivative at the peak (the curvature at the optimum)~\cite{Fisher}.

We now expand the posterior about its value at  $\theta=\MLEn$, by noting that $\theta-\MLEn$ is $O(n^{-\frac{1}{2}})$ (i.e. the width of the peak of $\Like_{n}(\theta)$ about the optimum scales as $\frac{1}{\sqrt{n}}$). We now write~\cite{Barndorff}:
\[
\pi^{*}_{n}(t) \doteq C_{n}^{-1}\exp\left[l_{n}(\theta) - l_{n}(\MLEn)\right]\pi(\theta),
\]
where $C_{n}$ is the normalisation term. The Taylor expansion for the log-likelihood is:
\begin{align}
l_{n}(\theta)-l_{n}(\MLEn) &= \frac{1}{2}(\theta-\MLEn)^{2}l_{n}'' + \frac{1}{6}(\theta-\MLEn)^{3}l_{n}''' + O(n^{-1}),
\end{align}
where we have used prime notation to denote multiple derivatives, and remembering that since we are expanding about the optimum, $l_{n}'\equiv 0$. Note also that since $l_{n}$ is trivially $O(n)$, we have to include the third powers to get a term of $O(n^{-\frac{1}{2}})$. We also expand the prior:
\[
\pi(\theta) = \pi(\MLEn)\left[1 + (\theta-\MLEn)\left(\frac{\pi'}{\pi}\right) + O(n^{-1})\right],
\]
which is just the first, linear, correction to a constant prior.

Substituting in for $(\theta-\MLEn)$ in terms of $t$, after some algebra we find:
\begin{align}
\pi^{*}_{n}(t) &= \frac{1}{\sqrt{2\pi}}\exp\left(-\frac{1}{2}t^{2}\right)\left[1 + \frac{t}{(n\hat{I}_{n})^{\frac{1}{2}}}\left(\frac{\pi'}{\pi}\right) + \frac{t^{3}l'''_{n}}{6(n\hat{I}_{n})^{\frac{3}{2}}}\right] \nonumber \\
 &+ O(n^{-1}).
\end{align}
We note that this expression is in agreement with the related expressions given by Welch and Peers~\cite{Welch1963}, although their method is more general, since they use the full moment generating function. The zeroth order piece of the posterior is a centred, unit Gaussian, which shows that we correctly scaled and shifted the parameter $\theta \rightarrow t$.  The first-order corrections are both odd polynomials in $t$, hence the normalization is just the usual term $\frac{1}{\sqrt{2\pi}}$ for a Gaussian, with no corrections required at this order\footnote{The pi in the normalization should not be confused with the function $\pi(\cdot)$ used for the prior.}. The full result says that the first correction to the zeroth-order Gaussian comes from two terms, the third derivative of the Likelihood at the optimum (which gives the amount to which the Likelihood is not symmetric about that optimum), and the term which shows to what extent the prior is not symmetric about the optimum (that is, if $\pi' \ne 0$).

We might want to try getting these two terms to cancel, and hence have a posterior that is Gaussian to $O(n^{-1})$. But we cannot do this algebraically in $t$, since the expansion in powers of $n$ mixes powers of $t$, so that here we have both linear and cubic terms in $t$ of the same order in $n$. However, given the statements above about the symmetry of the Likelihood, and the symmetry of the prior, we can instead require that the posterior should also be `symmetric', or at least centred. That is, if we require that the expectation value of $t$ under the posterior vanishes to first order:
\begin{equation}
E_{\pi^{*}_{n}}\left[t\right] \doteq \int dt \: t\; \pi^{*}_{n}(t) = 0 + O(n^{-1}).
\end{equation}
The integrals can be computed, since we just need the expectation values of powers under a unit Gaussian. These are given by the formula:
\[
E_{N(0,1)}\left[ t^{2m} \right] = \frac{(-1)^{m}}{\sqrt{2}}\left.\left(\frac{d}{d\alpha}\right)^{m} \frac{1}{\sqrt{\alpha}}\right|_{\alpha = \frac{1}{2}},
\]
where $N(\mu,\sigma^{2})$ is a Gaussian distribution of mean $\mu$ and variance $\sigma^{2}$. We hence find:
\begin{equation}
E_{\pi^{*}_{n}}\left[t\right] = \frac{1}{\sqrt{n}\hat{I}_{n}^{\frac{3}{2}}}\left[
\hat{I}_{n} \left(\frac{\pi'}{\pi}\right) + \frac{1}{2n}\frac{d^{3}l_{n}}{d\theta^{3}}
\right]_{\theta=\MLEn} + O(n^{-1}).
\end{equation}
The posterior is hence centred to first-order if:
\[
\hat{I}_{n} \left(\frac{\pi'}{\pi}\right) + \frac{1}{2n}\frac{d^{3}l_{n}}{d\theta^{3}} = 0.
\]
Using the definition of $\hat{I}_{n}$ from (\ref{eqn:FINdef}), this can be rearranged to give the differential equation:
\begin{equation}
\hat{I}_{n}\pi' = \frac{\pi}{2}\frac{d}{d\theta}\hat{I}_{n} \:\: \Rightarrow \:\:
\frac{d}{d\theta}\left[\pi(\theta)\hat{I}_{n}^{-\frac{1}{2}}(\theta)\right] = 0.
\end{equation}
This differential equation corresponds to equations (29) \& (30) of Welch and Peers~\cite{Welch1963}, and the equation for a first-order matching prior from the probabilistic-matching priors literature (e.g., see Ghosh~\cite{Ghosh2011} Eqn.~(4.3)). The solution of this equation is:
\[
\pi(\theta) \propto \hat{I}_{n}^{\frac{1}{2}}(\theta).
\]
which is just the Jeffreys General Rule prior~\cite{jeffreys1946invariant}. For a vector of parameters $\boldsymbol{\theta}$, the analogous Jeffreys prior (ignoring scaling and shifts) can be taken as:
\begin{equation}
\pi(\boldsymbol{\theta}) \propto \det(\mathbf{I}(\boldsymbol{\theta}))^{1/2}, \label{eqn:JeffDef}
\end{equation}
with the elements of the Fisher Information \emph{matrix} being defined as
\begin{equation}
I_{ij}(\boldsymbol{\theta}) \doteq  E\left[ -\frac{\partial^{2}} {\partial \theta_i \partial \theta_j}\logL \right]\label{eqn:FIMdef}
\end{equation}
These terms are otherwise known as the terms of the inverse parameter covariances $C_{ij}^{-1}$ ~\cite{Numerical} and can be recognised as
the standard approach for representation and estimation of parameter uncertainty.

\subsubsection{An Alternative Way to Understand Parameter Uncertainty}
Use of equation (10) assumes we are working in a regime where $n$ can be taken to be
large so that the uncertainty associated with parameters converges to a multivariate
Gaussian (Normal) distribution. This is generally \emph{not} the case for ANNs.
Difficulties of tractability also arise when applying equation (9) to evaluate
equation (2), both with the practicalities of computing second derivatives and also
the use of an expectation value (generally ignored).
Although it has been suggested that second derivatives should be computable on an ANN via a simple extension to the back-propagation training algorithm~\cite{Bishop}, the authors still know of no general method.
However, the na\"{\i}ve derivation of the simple Jeffreys prior above gives us the link between the Bayesian definition of Jeffreys invariant priors and the Frequentist literature on probability matching priors~(see \cite{Ghosh2011} and \cite{Datta}),
and indicates how we can progress.

It is important to note that the degrees of freedom we are manipulating here by defining a prior over the parameter(s) is our freedom to reparameterise our original family of model pdfs. A Likelihood or an integrated Likelihood is not a pdf or a probability. Under a redefinition of parameter $\theta \mapsto \omega(\theta)$, we have that:
\begin{equation}
\widetilde{\Like}_{n}(\omega(\theta)) \equiv \Like_{n}(\theta) \:\: \mbox{and:} \:\: \widetilde{\Like}_{n}(\omega(\theta))d\omega \equiv \Like_{n}(\theta)\left(\frac{d\omega}{d\theta}\right)d\theta
\equiv \widetilde{\Like}_{n}(\omega(\theta))\left(\frac{d\omega}{d\theta}\right)d\theta, \label{eqn:Ltildedef}
\end{equation}
where $\widetilde{\Like}_{n}$ is our new Likelihood function under reparameterisation. In particular, this means that the ordering of Likelihood values is preserved, hence $\hat{\omega}_{n} \equiv \omega(\MLEn)$ and the optimum Likelihood remains the optimum.

We hence see that the derivative $\frac{d\omega}{d\theta}$ of the reparameterisation function takes the place of the Bayesian prior $\pi(\theta)$, and the mapped Likelihood function $\widetilde{\Like}_{n}(\omega(\theta))$ generated from the original Likelihood function replaces the posterior $\pi^{*}_{n}(\theta)$~\cite{Welch1963} as generated from the Likelihood function. The requirement that a prior pdf is non-negative becomes the requirement that our reparameterisation function $\omega(\theta)$ is monotonically non-decreasing (that is, $\frac{d\omega}{d\theta}\ge 0 $).
The use of a Jeffreys prior can be considered as only the start of an iterative process, for which the Gaussian mapped parameter is ultimately the result (see Appendix~\ref{app1})
\footnote{An iterative sequence of invariant priors, starting from the Jeffreys prior, was also investigated by Dowe. See~\cite{Dowe}, \S7.1, page 953 for an example involving the multinomial distribution.}.

Rather than applying the Jeffreys prior/mapping itself, singly or iteratively
, we instead chose to map a suitable portion of the Likelihood function directly to a Gaussian.
We hence will require that:
\[
\frac{\Like_{n}(\theta)}{\Like_{n}(\MLEn)} = \frac{\widetilde{\Like}_{n}(\omega)}{\widetilde{\Like}_{n}(\hat{\omega})} = \exp \left(-\frac{1}{2}(\omega-\hat{\omega})^{2}\right) \:\: \Rightarrow \:\: \ln\left(\frac{\Like_{n}(\theta)}{\Like_{n}(\MLEn)}\right) = -\frac{1}{2}(\omega-\hat{\omega})^{2}.
\]
Therefore, if we centre the MLE such that $\hat{\omega} \equiv \omega(\MLEn) = 0$ then:
\begin{equation}
\omega(\theta) = \sign (\theta-\MLEn)\sqrt{-2\ln (\Like_{n}(\theta)/\Like_{n}(\MLEn))} \equiv z(\theta),\label{eqn:zdef}
\end{equation}
which is just the function $z(\cdot)$, defined as the signed square-root of the log-likelihood ratio statistic~\cite{Jensen,DiCiccio}.
It has been shown previously that hypothesis tests constructed for
$z(\theta)$ are
rectangular ~\cite{Welch1963} and honest ~\cite{Dawid}, as required for
practical use.
The idea can also be seen to be consistent with the work of Cram\'{e}r and Rao regarding the minimum variance bound~(MVB), in the sense that the MVB is saturated when the Likelihood function is exactly Gaussian with a known mean, since efficient estimators for the variance exist in this case~(e.g., see Cram\'{e}r~\cite{Cramer}, Chapter 32).

This general approach was known about at least as far back as Anscombe\footnote{Who also said~\cite{Anscombe1964}, ``\emph{Typically it is the evidence from a small body of data (often corresponding to a non-normal Likelihood function) that is difficult to grasp precisely.}"} in 1964~\cite{Anscombe1964, Hougaard}.
It can also be related to the more familiar case of the Fisher $z$-transformation{\footnote{For the specific case of sample correlation coefficients, the highly-skewed nature of the sampling distribution, even for large sample sizes, made the standard correlation coefficient unsuitable when it came to assessing the accuracy of observed correlations. Fisher showed that a simple transformation based on the hyperbolic tangent reduced these curves to close approximations to the normal distribution, with a variance that is stable over different values of the true correlation~\cite{FisherZ1915,FisherZ1921}.}}, which
Winterbottom~\cite{Winterbottom} shows can be derived by first requiring that it reduces skewness, and that, after bias correction, is \emph{both} normalising and variance-stabilising.

In conclusion, the Jeffreys prior can be derived as an approximation to the
process of mapping the original Likelihood function onto a Gaussian.
Frequentists would claim that this is what the Jeffreys prior is doing,
whilst Bayesians might claim that equation (9) defines the underlying
principle.
What we explain below is that it does not matter which of these
explanations you personally prefer, the consequences will turn out to be the same.

\subsubsection{Theory Summary}
Jeffreys derived his priors in order to obtain consistency under
parameter transformation, which can
be considered a scientific requirement.
In a Frequentist sense, the Jeffreys prior can be derived as a density
scaling which approximates the mapping of a parameter to achieve a
Gaussian Likelihood function (see Eqns.~(\ref{eqn:FINdef}) to (\ref{eqn:JeffDef})).
It should be noted that such a mapping is also consistent under parameter transformation.

Generally when performing Likelihood estimation of parameters, we rely to some extent
on the central limit theorem to ensure that for sufficiently large quantities of data
the Likelihood function around the optima will be approximately multi-dimensional Gaussian.
Under these circumstances the derivative terms
and the correlations between them,
can be modelled using a covariance matrix determined from the Minimum Variance Bound (MVB)
in the usual way ~(\ref{eqn:FIMdef}). However, for highly asymmetric likelihood functions
proportionately more data will be needed.
For highly non-linear systems, and in circumstances
where the curse of dimensionality reduces the effective data quantity,
we can not expect this justification to hold. We will show
below that this is the circumstance which we encounter for ANNs, but first
we start with two simpler problems in statistical estimation which
illustrate this.

Rather than using Jeffreys' approximation, if we directly map a parameter
with $z(\theta)$ (\ref{eqn:zdef}), {\bf the Jeffreys
prior density $\pi(z(\theta))$ is not only exact but uniform},
due to origins of Eqn.~(\ref{eqn:JeffDef}). In a Bayesian sense, you may not accept
this origin for Jeffreys priors and would prefer to simply
accept Eqn.~(\ref{eqn:JeffDef}) as already exact. Either way, for this special
definition of a parameter the Likelihood
function can be directly interpreted as a parameter density (\ref{eqn:postdef}).
Under this scheme the Bayesian and Frequentist approaches are directly
comparable and we achieve a form of synthesis,
in all respects except the interpretation of $\pi(\theta)$
as a probability{\footnote{Bayesians have already noted that Jeffreys
priors are often not consistent with Kolmogorov's axioms and therefore ``improper''.}}.

Now that we have this understanding of the relationship between
approaches, it gives us new analysis options. We can exploit either: the observation that the
uninformative prior for the original
parameter is the derivative of $z(\theta)$ (see Approach I below) , or that that the prior for this
mapped parameter is uniform (see Approach II below).
Both insights allow us to avoid the need to evaluate equation (9) but
still compute equation (2).

\subsection{Parameter Uncertainty: Approach I, Binomial and Chi-square}

In what follows, we use the signed square-root of the log likelihood ratio defined above as our mapping function. We then estimate the distribution over a parameter for use in assessment of future computational uncertainty (systematic or `epistemic' errors).  That is:
\begin{equation}
p(\theta|\mathbf{X}) \propto  \Like(\theta,\mathbf{X})~ |\frac{d z}{d \theta}|,\label{eqn:zpriordef}
\end{equation}
where we now use the $p(\theta|\mathbf{X})$ notation rather than the posterior $\pi^{*}(\theta,\mathbf{X})$ to make it clear that we are using a \emph{specific} mapping of parameter space rather than an explicit Bayesian prior on parameter space, or a general mapped Likelihood function.
We emphasise at this point that, following basic local argument,
this expression is expected to be exact under either a
Frequentist or Bayesian interpretation and applicable to arbitrary likelihood functions, whilst use of equation (9) is not.

\subsubsection{Example: The Estimation of a Variance from a Sample with Known Mean}
We consider a very small sample $n$ of Gaussian i.i.d. data $\{X_{i}:i=1,\ldots n\}$, where the mean $\mu$ is known, and the model is parameterised by the variance thus:
\[
p(x|v) = \frac{1}{\sqrt{2\pi v}}\exp\left(-\frac{1}{2v}(x-\mu)^{2}\right).
\]
The variance parameter has been chosen as an example since it gives a highly skewed Likelihood function, and also to illustrate the non-unique nature of Jeffreys various priors~\cite{Kass}. The Likelihood function is:
\[
\Like_{n}(v) = \frac{1}{(2\pi v)^{\frac{n}{2}}}\exp\left(-\frac{1}{2v}\sum\limits_{i=1}^{n}(X_{i}-\mu)^{2}\right).
\]
We are asked to determine the uncertainty associated with a MLE of variance $\hat{v}$, where\footnote{Note that this is an unbiased estimate since we are using the known mean, rather than the sample mean.}:
\[
\hat{v} \doteq \frac{1}{n} \sum\limits_{i=1}^{n} (X_i - \mu)^2,
\]
and hence the corresponding log-likelihood function is:
\[
- \logL_{n}(v) = - l_{n}(v) = \frac{n}{2}\left[  \frac{\hat{v}}{v} +  \ln(2 \pi  v)\right].
\]
What we can observe for this system is that once $\hat{v}$ is specified, the Likelihood can be written in a scale-invariant form:
\[
\Like_{n}(v,\hat{v}) \propto \frac{1}{(2\pi)^{\frac{n}{2}}} \cdot \left(\frac{\hat{v}}{v}\right)^{\frac{n}{2}}
\exp \left[ -\frac{n}{2}\left(\frac{\hat{v}}{v}\right)\right] \propto \left(\frac{\hat{v}}{v}\right)^{\frac{n}{2}}\exp \left[ -\frac{n}{2}\left(\frac{\hat{v}}{v}\right)\right].
\]
We can therefore, without loss of generality, restrict ourselves to
consideration of the uncertainty associated with $\hat{v}=1$.  We can now compare the theoretical predictions from Bayesian and Frequentist approaches. For this particular example there are two Jeffreys priors in the literature~~\cite{Kass}~(see Table \ref{tab:dists}). The first is the non-location or scale invariant prior (see Jeffreys~\cite{jeffreysbook}, \S3.1), which gives $\frac{d\sigma}{\sigma} \propto \frac{d(\sigma^{2})}{\sigma^{2}} = \frac{dv}{v}$, whereas the second is the Jeffreys General Rule prior~\cite{jeffreys1946invariant}\footnote{To be precise, this is the General Rule prior when you take the model to be that with two parameters, $\Like(\sigma,\mu,\mathbf{X})$, where you compute the determinant of the Fisher Information matrix to obtain the prior. The fact that the General Rule itself gives a \emph{different} answer if you fix one parameter and compute just the Fisher Information function is the specific example considered by Jeffreys in the 1961 edition of~\cite{jeffreysbook}, see \S3.10, page 182.}, where $\frac{d\sigma}{\sigma^{2}} \propto \frac{dv}{v^{\frac{3}{2}}}$.

\begin{table}
\begin{center}
\begin{tabular}{|l|l|l|}
\hline
  {\bf Approach} & {\bf prior $\pi(\theta)$} & {\bf conditional density $p(\theta|x)$}  \\
\hline
&Chi-square variance  $v$& \\
\hline
  Jeffreys General Rule & $\pi(v)~=~1/ v^{3/2}$ & $ \Like(v)/v^{3/2}$ \\
  Jeffreys Non-location Rule & $\pi(v)~=~1/v$ & $ \Like(v)/v$  \\
  Frequentist  & $\pi(v) ~=~ 1. | \partial z(v) / \partial v  |$ & $ \Like(v) | \partial z(v) / \partial v  |$  \\
\hline
&Binomial $p$ & \\
\hline
  Jeffreys General Rule & $\pi(p)~=~1/\sqrt{p(1-p)}$ & $\Like(p)/\sqrt{p(1-p)}$  \\
  Frequentist  & $\pi(p) ~=~  1.| \partial z(p) / \partial p  | $& $  \Like(z(p)) | \partial z(p) / \partial p  |$  \\
\hline
\end{tabular}
\caption{Example: Variance estimate distribution using uninformative priors. Note that
the frequentist theory interprets the Bayesian prior as a product of two terms,
a uniform density over the mapped variable $z$ (consistent with the Fisher information being constant), which can be seen
as the true ``prior'', and the differential term needed to conserve probability mass
under variable transformation away from $z$. Also $\Like(z(\theta))~=~ \Like(\theta)$. }
\label{tab:dists}
\end{center}
\end{table}

\begin{figure}[t!]
\begin{center}
\subfigure[]
{
\includegraphics[scale = 0.5]{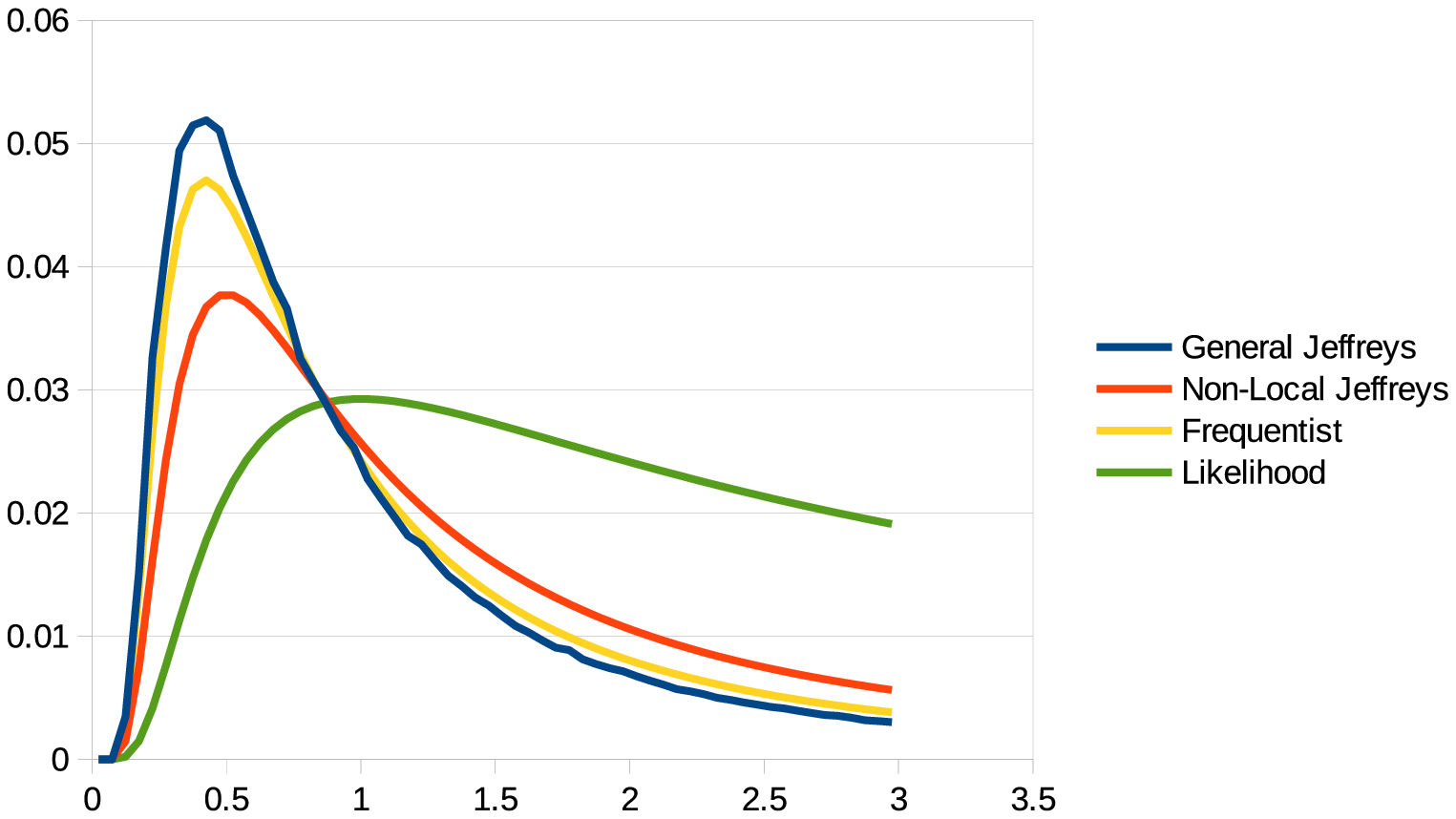}
}
\subfigure[]
{
\includegraphics[scale = 0.5]{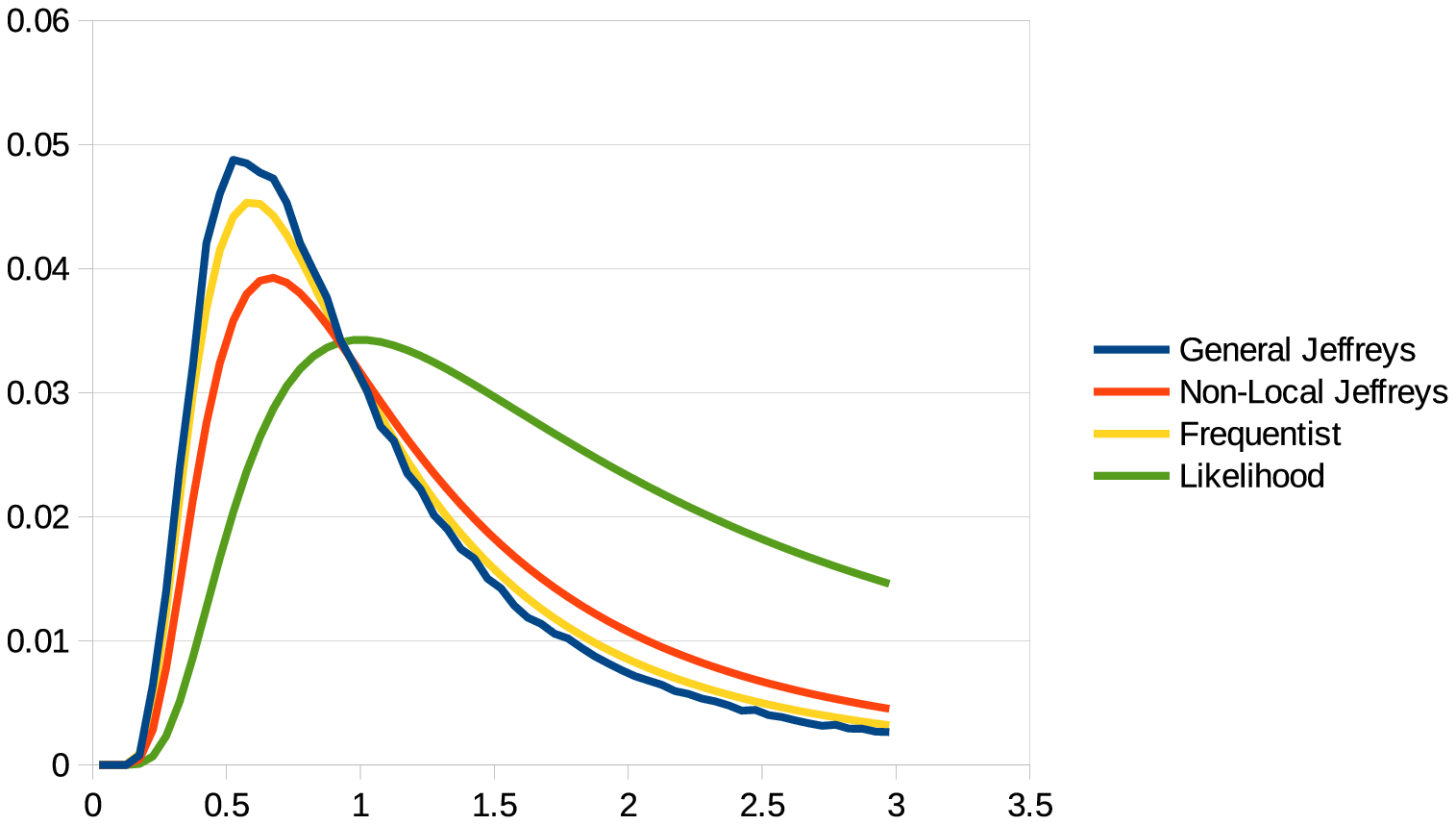}
}
\caption{\small{
Theoretical predictions of the distributions of
variance estimation uncertainty for (a) a sample of n=2, with $\hat{v}=1$ and
(b) a sample of n=4 , with $\hat{v}=1$. Note that for Jeffreys priors of the form $1/v^{a}$, the peak occurs at $v_{\mathrm{peak}} = n/(n+ 2a)$.
}
}\label{fig:Ex1}
\end{center}
\end{figure}

The Frequentist approach, though a different solution to either of the previous two, is also invariant under monotonic non-linear remapping of $v$. It uses the mapping generated by $z(v)$ as defined by Eqn.~(\ref{eqn:zdef}). In this case:
\[
\Like_{n}(v) = \Like_{n}(v,1) \propto \frac{1}{v^{\frac{n}{2}}} \exp \left[-\frac{n}{2v}\right]
\:\: \Rightarrow \:\: z(v) = n^{\frac{1}{2}}\sign(v-1)\sqrt{\ln v + \frac{1}{v}-1},
\]
and it is straightforward to check that the Likelihood becomes a Gaussian
when rewritten in terms of $z(v)$. The derivative term, which would be interpreted as a prior under a Bayesian framework, is instead the term needed to conserve probability mass under parameter transformations $z \mapsto v$, as we move away from a Gaussian Likelihood function. These definitions lead to the distributions over $v$ shown in Figure~\ref{fig:Ex1}. Note that although $\widetilde{\Like}(z)$ plotted against $z$ is a Gaussian by definition, for comparison with the other approaches we instead have to consider \mbox{$\Like(v)\pi(v) \equiv \widetilde{\Like}(z(v))\cdot 1 \cdot | dz(v)/dv |$} plotted as a function of $v$ (see Eqn.~(\ref{eqn:Ltildedef}) for details).

Note that for a Jeffreys prior of the form $1/v^{a}$, the peak of the plots will lie at a value of $v_{\mathrm{peak}} = n/(n+2a)$. For small $n$, this will not be close to the theoretical expectation value of $\hat{v}=1$. For the frequentist plots, note that although the peak of $\Like(v)$ will lie exactly at $\hat{v}=1$, the same is \emph{not} true for the plot of $\Like(v) |dz(v)/dv|$.


The general observation of these curves is that they are all quite similar, certainly
to the level of difference observed between different Jeffreys rules. The Frequentist
Monte-Carlo distribution is shown for comparison. DiCiccio and Martin~\cite{DiCiccio} observe that
this approach gives ``near perfect coverage'' in hypothesis tests, i.e. the estimated
distribution is quantitatively valid, as required for a valid Frequentist theory.

\subsubsection{Example: Estimation of Binomial Probability}

Binomial statistics can be used to model the estimation of a probability using
sample data, i.e. $n$ from $N$ samples, $\hat{p} = n/N$.  When using ANNs,
the approximation of outputs as conditional probabilities
($o~\approx~P(C|{\bf \mathbf{X}})$) is normally explained as a local estimation
of the proportional ratio of classes observed in the sample $\mathbf{X}$ ~\cite{Lippmanb}.
This can be achieved using either least-squares or cross entropy cost
functions, although the correct Likelihood function for this is still better
understood in statistical terms as binomial statistics.
However, the potential for large dimensional input vectors leads to the well known ``curse of dimensionality",
and in some places in the data space, outputs have to
be estimated with very small sample sizes, perhaps even only one.
In these situations, the Likelihood estimate of $P(C|{\bf \mathbf{X}})$ may be high, but the
uncertainty in this calculation is not reflected in this value. Under these
circumstances it is better to think in terms of the expectation of the output,
as this quantifies the uncertainties due to sample statistics.

\begin{figure}[!b]
\begin{center}
\subfigure[]
{
\includegraphics[scale = 0.5]{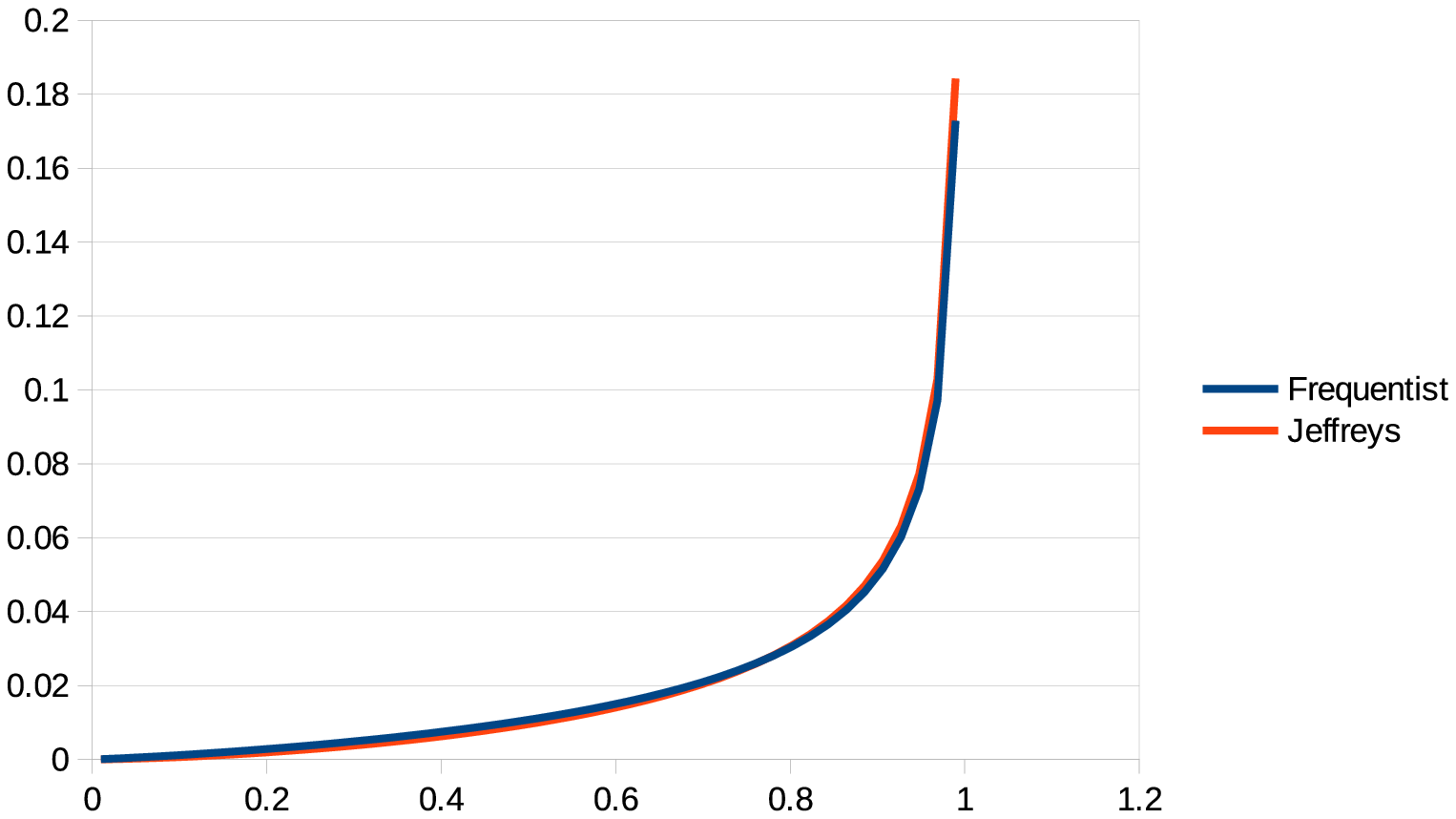}
}
\subfigure[]
{
\includegraphics[scale = 0.5]{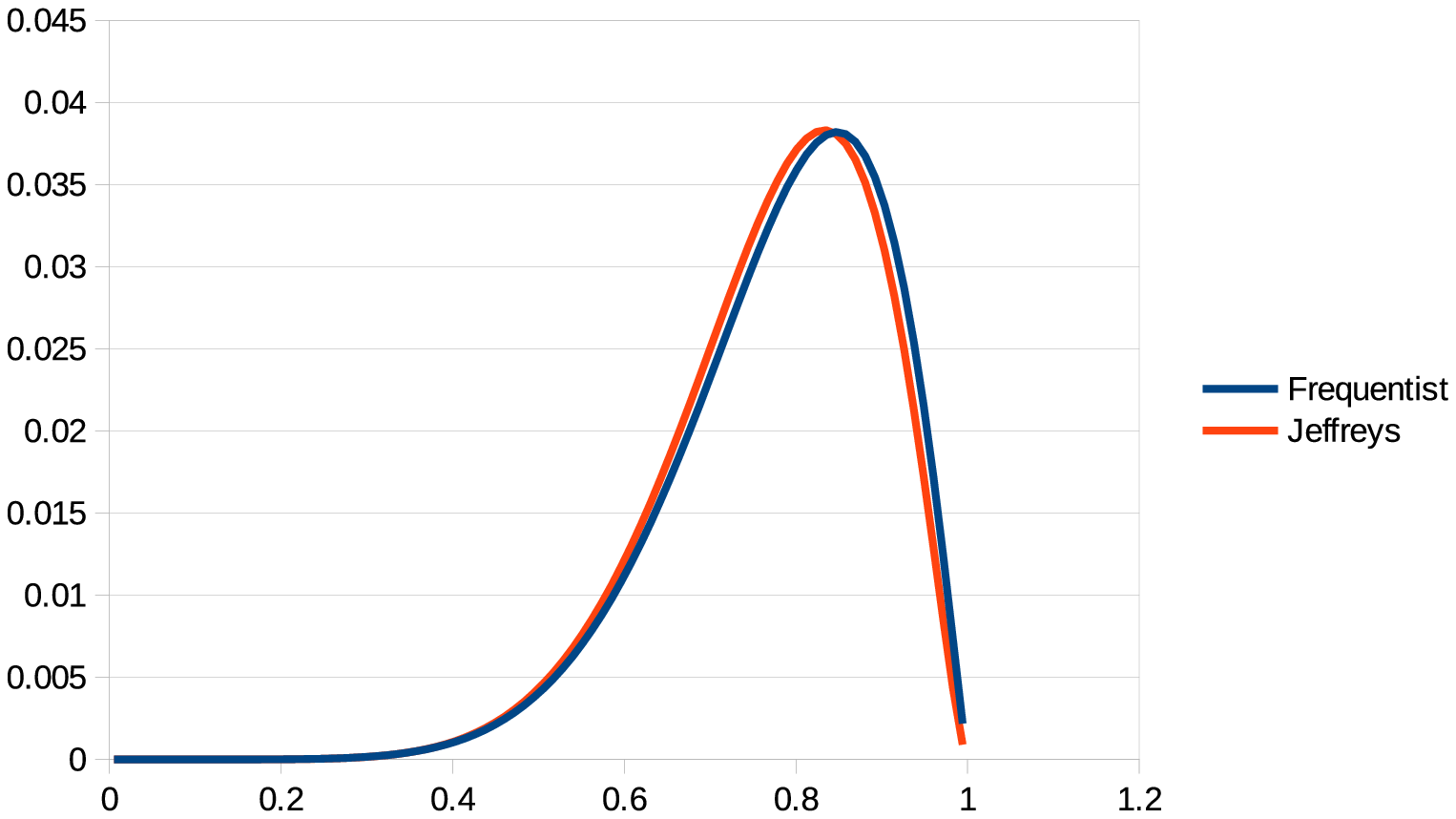}
}
\caption{\small{
Theoretical predictions for the distributions of
binomial probability ($p$) estimation uncertainty for (a) a sample of N=2, with $\hat{p}=1$,
and (b) a sample of N=10 with $\hat{p}=0.8$.
}
}\label{fig:Ex2}

\end{center}
\end{figure}

For this example we therefore examine the task of estimating $o ~=~n/N$ given
very small sample sizes, and compute the distribution over this value using Bayesian
and Frequentist approaches. Hence now the data we have is the value of $n$ observed, and the relevant parameter to be determined is the related probability $p$. This then gives a Likelihood for this system as:
\begin{equation}
\Like(p,n) = p^n (1-p)^{N-n},
\end{equation}
where as expected, the MLE of $p$ is given by $\hat{p} = n/N$. The Jeffreys General rule prior is \mbox{$\pi(p) \propto  \sqrt{\left(\frac{N}{p(1-p)}\right)}$},
and the (Frequentist) Gaussian mapping function is:
\begin{equation}
z(p) =   \pm \sqrt{-2\ln \left(\frac{p^n (1-p)^{N-n}}{\hat{p}^n (1-\hat{p})^{N-n}}\right)},
\end{equation}
and the sign is determined by the sign of $(p-\hat{p})$. Note that $z(p)$ is correctly defined so that $z(\hat{p}) = 0 = \hat{z}$.
The mathematical summary is given in Table~\ref{tab:dists}, and the calculated uncertainty for $\hat{p} = n/N = 2/2$ and $\hat{p} = 8/10$ are shown in Figure ~\ref{fig:Ex2}.
We can see that the computed distributions from both theoretical approaches are nearly identical.

\subsubsection{Summary}
The Frequentist and Bayesian approaches to the calculation of
uncertainty over parameter determined using Likelihood are very similar,
even for small sample situations where the difference might be expected to be at there largest.
Given the aims of this work, to summarise the uncertainty associated with
likelihood estimation, it is the Frequentist theory, based upon making quantitative
distributions ``honest'' for use in hypothesis test construction, which most
directly addresses our needs. The Bayesian approach omits the Frequentist
axiom by choice (i.e. any need to conform to data samples). It is
an approximation which is derived on
the basis of more general mathematical considerations. It therefore seems to us
very natural to use the Frequentist approach as the basis for the modelling
of uncertainty in ANNs.

\newpage
\subsection{Parameter Uncertainty: Approach II, ANNs}

Having shown how the theory deals with standard low sample statistical problems,
we now wish to apply it to parameter uncertainty in ANNs. Approach I is not
entirly practical for this.
However, based upon the principle outlined above, by replacing the original parameter $\theta_w$
with the remapped variable $z(\theta_w)$ we replace $\pi(\theta_w)$ with a constant $\pi(z(\theta_w))$.
Then the likelihood function  also describes the (Gaussian) uncertainty over the parameter.
However, there is still a problem. Strictly the mapping needs to be applied to the
multi-dimensional parameter space, i.e.  ${\bf z}$ is a vector function.
What we propose in this work is that we apply separate transforms
$z(\theta_w)$ to each
of the $W$ network parameters.
In effect we assume the ``prior'' can be written as
\begin{equation}
\pi({\bf z}(\Theta)) ~\approx~ \prod\limits_{w=1}^W \pi(z(\theta_w)) = \mathrm{constant}.
\end{equation}
So that by replacing each parameter $\theta_w$ with $z(\theta_w)$,
this  will make the joint parameter distribution approximate a
multi-dimensional Gaussian. If nothing else we would
hope that this will help accelerate the
effect of the ``central limit'' process (see section 1.1.3).
Then the original likelihood function can also be approximated using
\begin{equation}
-\ln[L({\bf z}(\Theta))] \approx  -\ln[L({\bf z}_0)]  +   {\bf z}^T ~C_{\bf z}^{-1}  {\bf z}, \label{eqn:zL}
\end{equation}
and as the ``prior'' terms are constant
\begin{equation}
p({\bf z}| {\bf X}) \propto \exp [-  {\bf z}^T ~C_{\bf z}^{-1}  {\bf z}]. \label{eqn:zprior}
\end{equation}
We can then use the inverse covariance  ($C_{\bf z}^{-1}$  )
to approximate the uncertainty
in the parameters arising due to the training sample.
The second part of this paper now details how this approach was
implemented and tested for a real world clinical decision support system and
the insights gained.

\section{Methods}

\subsection{Data Acquisition and Preparation}

The dataset chosen to illustrate the estimation of ANN output uncertainty
is the clinical task of early diagnosis of dementing diseases, where a safe interpretation
would be considered essential for ethically responsible patient management.

\begin{center}
  \begin{tabular}{ccccc}
    \hline
  \label{table6}
    Diagnosis & Norm. & Alz. & F.T.D. & Vas.D. \\
    \hline
    Age ~ (sd) & 64.2 ~(7.7) & 61.3 ~(6.4) &60.6 ~(0.2) & 67.6 ~ (5.9)   \\
    duration ~(sd) & - & 3.4 ~(1.6) & 3.6 ~(3.1) & 2.3 ~(2.1) \\
    \hline
  \end{tabular}
\newline

Table 2. Demographic make up of the sample.
\end{center}

The subjects comprised 19 patients with frontotemporal
dementia, 18 with Alzheimer's disease, 11 with vascular dementia and 9
normal controls. Their age distribution and duration of illness are
shown in table 2. All patients were referrals to a specialist
diagnostic dementia clinic, and had undergone comprehensive neurological
and neuro-psychological assessments as part of their diagnostic
evaluation. Patients with frontotemporal dementia and Alzheimer's disease
fulfilled currently accepted clinical diagnostic criteria for those
conditions ~\cite{ Lund, Neary, McKhann} and were free from significant
risk factors for cerebrovascular disease (Hachinski scale $<$ 4)
~\cite{Hachinski}. Patients with vascular dementia all had high risk factors for
vascular disease, with Hachinski scale scores $ > $ 7.  Patients exhibited the
characteristic pattern of dementia associated with their clinical diagnosis
~\cite{ Snowden}.
All patients had been followed up for
$1~-~3.5$ years. The clinical diagnosis was therefore confirmed by the
evolution of the illness.
Individuals were excluded if diagnosis
of the form of dementia was equivocal or if the clinical pattern
suggested mixed aetiology.

All subjects were scanned using a Phillips 1.5 Tesla ACS-NT
scanner with a PowerTrack 6000 gradient subsystem. The patients were
scanned using a birdcage head-coil receiver. CSF segmentation was
performed on coronal fast spin echo inversion recovery images (TR 6850 ms,
TE 18 ms, TI 300 ms, echo train length = 9 ).  Contiguous 3mm slices were
obtained throughout the brain with an in-plane resolution of  0.89 mm2
(matrix 256 x 204,  field of view 230mm x 184mm ). The details of the image
analysis are given in ~\cite{Thacker}.
The CSF volume measurements were obtained twice for each subject, giving a total
of 118 samples.

The purpose of using an ANN in this work is to map twelve volume measurements
of cerebral-spinal fluid (CSF) volume to four diagnostic categories ($C$): normal brain,
Altzheimers disease, frontotemporal dementia and vascular dementia.
CSF was selected for this work because it is easier to segment from MR images
than brain tissue, and gives a direct quantitative assessment of the change
in brain-volume over time, as the skull size remains fixed  after early adulthood.

However, some additional input variables are also needed to correctly
interpret CSF volume measurements, as total brain volumes vary in
size between individuals and the normal degree of atrophy varies as a function of age. Including age
and volume to the list of inputs gives a total of 14 along with
four output variables. If we try to directly train an ANN on the 118 examples
available there would not be enough data to constrain the free parameters.
In previous work ~\cite{Thacker}, this issue was addressed via application of prior knowledge
(see Discussion below) and use of a k-nearest neighbour classifier (kNN).
Here the kNN is replaced by an ANN.

\begin{figure}[ht]
\begin{center}
\subfigure[]
{
\includegraphics[scale = 0.5]{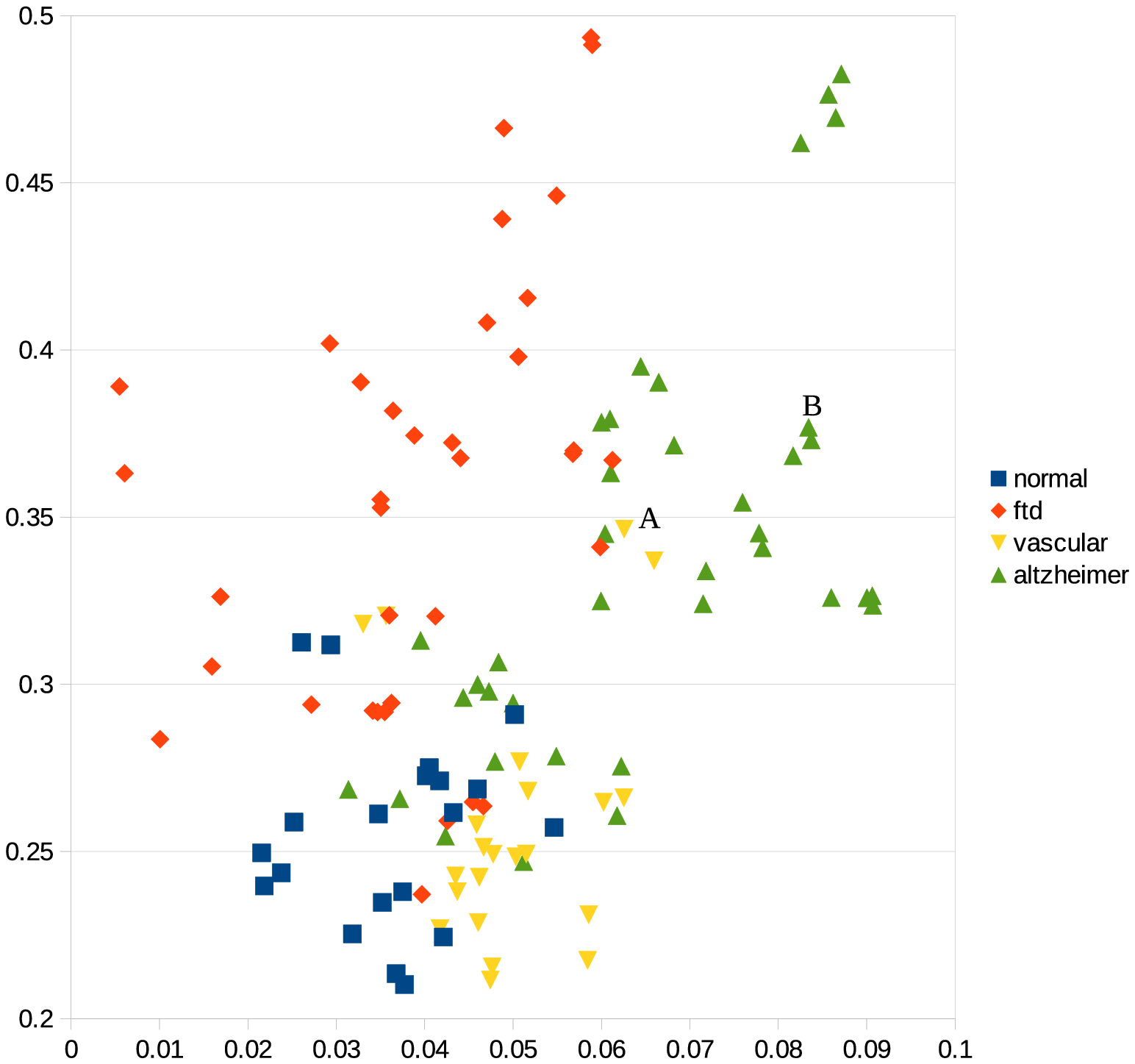}
}
\subfigure[]
{
\includegraphics[scale = 0.5]{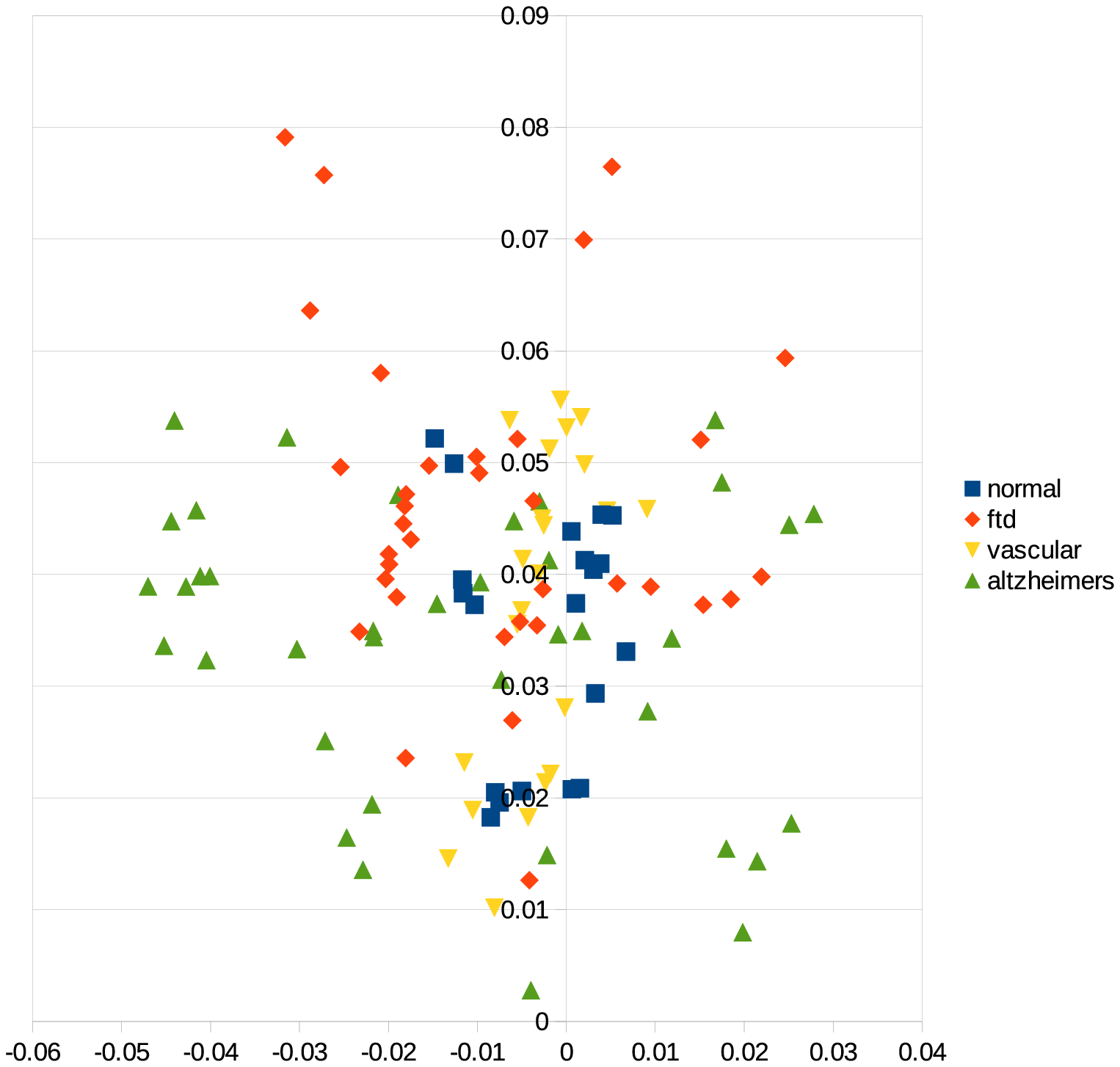}
}
\caption{\small{
(a) Graph of reduced variables $X_3$ vs $X_2$.
parameters.
(b)  Graph of reduced variables $X_1$ vs $X_4$.
}
}\label{fig:reduced}
\end{center}
\end{figure}

The initial twelve volume measurements were normalised to the volume of a rectangular bounding
box of the CSF. The normal subjects were used to determine a simple (and therefore stable,
low parameter) correction, which adjusted each CSF volume of normal subjects back
to a nominal age around 40. These same corrections were then applied to diseased data.
The typical expected behaviour of dementing diseases was used to further
reduce the number of dimensions from 12 down to 5 (see Discussion).
The final five variables can be represented as a five dimensional vector ${\bf X}$ which
is used as input to an ANN for purposes of estimating $P(C|{\bf X})$ (see Figure~\ref{fig:reduced}).

\subsection{ANN Architectures and Training}

A selection of ANNs were generated with 5 inputs and four outputs ($o_{nm}$)
with binary training targets $x_{nm}$ and
variable numbers of hidden units with either one or two hidden layers.
These networks were trained using a combination of Resilient Propagation ~\cite{RPROP}
and Conjugate Gradient Optimisation ~\cite{Numerical}, using purpose written software (ANSI C).
It was important for
later stages of the experiments that the cost function had well located optima $\Theta_o$.
This was generally achieved after 20,000 cost function evaluations for RPROP
followed by a further 1,000 evaluations using conjugate gradient optimisation.
This completed on a DELL Precision T7500 workstation, using only one of
the co-processors,  in under one minute.

The cost function used was cross entropy defined as
\begin{equation}
Q(\Theta)~=~ -2 \ln[L(\Theta, {\bf X})] ~=~ 2 \sum\limits_{m=1}^4 \sum\limits_{n=1}^N \left[ x_{nm} \ln (o_{nm}) ~+~ (1-x_{nm}) \ln (1-o_{nm})\right],
\label{eqn:crossent}
\end{equation}
which can be considered a Binomial Likelihood function evaluated on a continuum of
input samples ($x_{nm}$). The factor of 2 allows a direct relationship to be made
between this Likelihood and Chi-squared statistics, and is also needed when re-mapping
onto a Gaussian ~(\ref{eqn:zdef}). We believe this to be the quantitatively valid Likelihood
to use for this classification task. It is important that the cost function is
selected in this way, as all subsequent theory regarding output uncertainty estimation
is only correct if the Likelihood function is an honest model of the training data
uncertainty.

The alternative architectures were evaluated on the basis of their generalisation
performance using both leave-one-out cross validation and calculation
of the Akaike Information Criteria (${\bar Q(\Theta)}~+~ 2~ N_\Theta$) ~\cite{Akaike}.
For simplicity the cross-validation score was computed as a least-squares
difference between the network output and the training target.
This involved training each tested architecture 118 times and ${\bar Q(\Theta)}$
was defined as the final cost function average.
Following this two architectures were chosen for uncertainty estimation
tests.

The Gaussian re-mapping of parameters was achieved using simple inspection of the
cost function.
Given a selected ANN, the weight parameters ($\theta_w$ ) were first assessed
to determine the typical scale $\Delta_w$ which generated a change in one of
the symmetrised cost function $(Q(\Theta_0+\Delta)+Q(\Theta_0-\Delta))/2$.
This was done using an iterative process requiring no more than
10 evaluations of the cost function.
This allowed us to generate graphs of the asymmetry in the cost function
over a sensible change of $\theta_w$ prior to remapping. It also allowed us
to assess the effect of asymmetrical parameter re-mapping on the parameter covariance.

Starting from $\Delta_w$, the change in $Q(\Theta)$ was assessed for each parameter
to determine three points on either side of the minimum $\Theta_0$, which spanned a change
in the cost function of around 3. This could typically be achieved with
less than 20 evaluations of the cost function. These points were then used to define
a cubic spline approximation to the inverse  of Eqn.~(\ref{eqn:zdef}) ($\theta(z_w)$), which allows
a value of $\Theta_w$ to be computed which will give rise to a specific value of $z$,
such that
\begin{equation}
Q(\Theta_0 ~+~ \Theta(z_w))~-~ Q(\Theta_0) ~=~ z_w^2
\label{eqn:spline}
\end{equation}
The adequacy of this mapping was checked to see if values of $z_w^2$ could be achieved
to an accuracy of 0.1 up to a maximum of $z= \pm 5$, requiring a further 20
cost function evaluations. The upper and lower
limit of this valid range were stored for later use.

The remapped parameters $z_w$ define a log Likelihood function which is
quadratic, at least for changes in individual parameters $w$.
In order to avoid the need for an analytic calculation of the second derivatives
of the cost function,
and to get a better approximation to the general shape of the cost function,
the off-diagonal inverse covariance  for pairs of parameters ($wv$)
were then computed via inspection using
the relationship
\begin{equation}
C_{wv}^{-1} ~=~ \frac{1}{8 \Delta^2}{\sum_{z_w =\pm \Delta}\sum_{z_v=\pm \Delta} Q(\Theta_0 + \Theta(z_w) + \Theta(z_v)) }
\label{eqn:offdiag}
\end{equation}
at the expense of 4 more cost function evaluations.
Here, $\Delta$ is chosen to be the maximum allowable absolute values
determined by the valid limits $z_w$ and $z_v$.
It is worth noting that this calculation is the least squares estimate
of the parameters of a quadratic model
using four symmetrically placed evaluations, and is relatively insensitive to
the exact location of the optima.
The diagonal terms of this matrix are all 1 due to the definition of $z$.

The cost function could therefore be remapped and the inverse covariance
estimated for $\approx$ 30 parameters with typically 2,000 evaluations of the cost function,
i.e. equivalent to around 10\% of the original network optimisation.

\subsection{Evaluation of Output Uncertainty}

The uncertainties in ANN output for an individual input sample was evaluated
using a Markov-Chain Monte-Carlo (MCMC).

Gaussian distributed IID random variables ($\sigma =  0.2$) were
used to generate
random steps in $z_w$. These steps were then evaluated using
Eqn.~(\ref{eqn:zprior}) and every 50th update was output as a sample in order to reduce
sample correlation.
A hundred of these samples were used to generate instances of network parameters
using the vector of cubic spline approximations $\Theta({\bf z})$.
Values of $z_w$ generated outside of the valid range of the mapping were
rejected by setting the corresponding density estimate in the MCMC
step to zero.
Accepted variations in ${\bf z}$ were used to build up the distribution over
the output for the selected input sample.

Finally, in order to confirm the adequacy of the Gaussian mapping, the computed
Mahalanobis distance was stored along with true ANN cost function for later comparison.
The combined process typically involved 10,000 cost function evaluations,
but this could be reduced to as few as 100 if the Gaussian mapping comparison
was not needed.

\section{Results}

\begin{figure}[ht]
\begin{center}
\subfigure[]
{
\includegraphics[scale = 0.5]{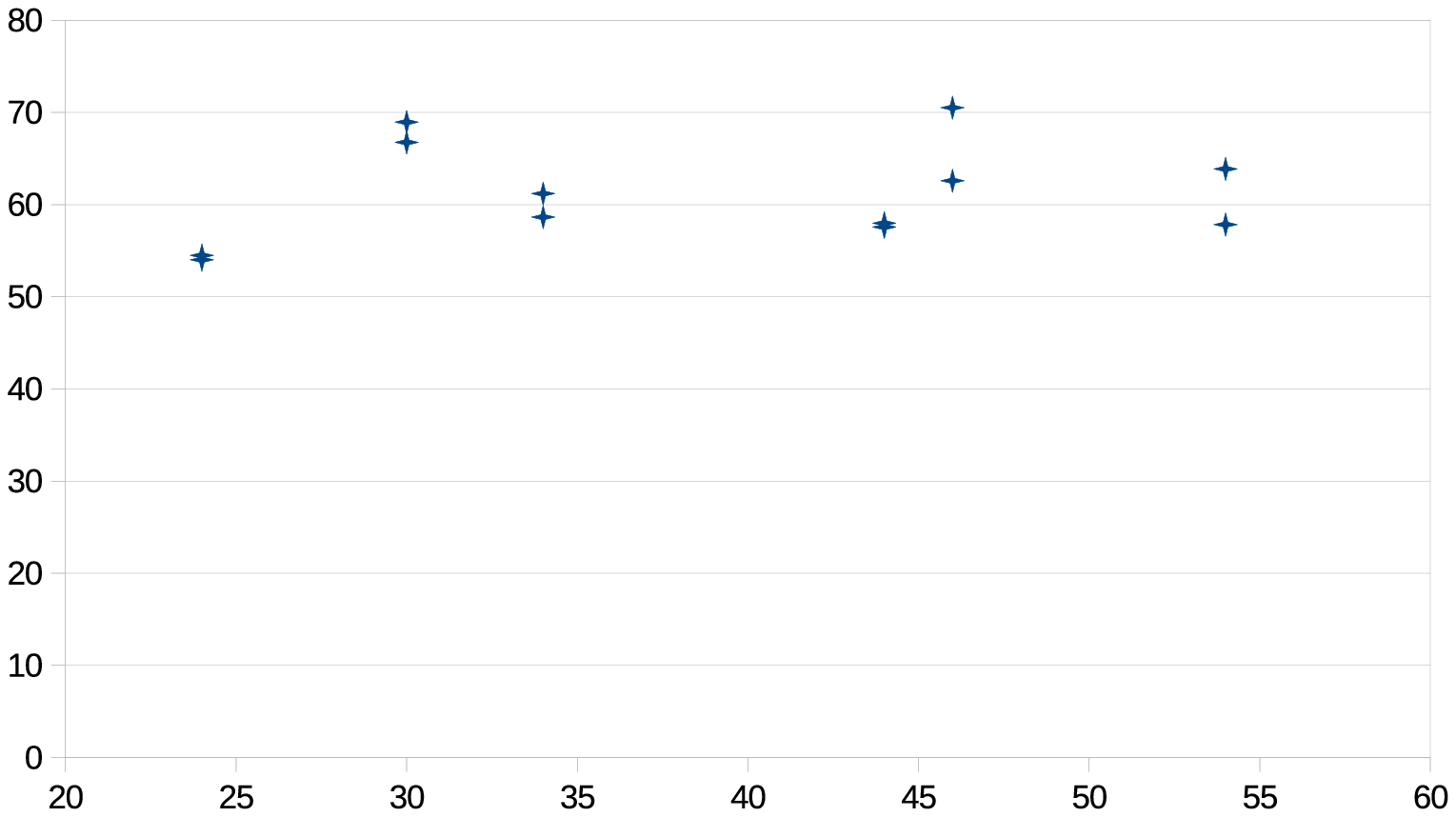}
}
\subfigure[]
{
\includegraphics[scale = 0.5]{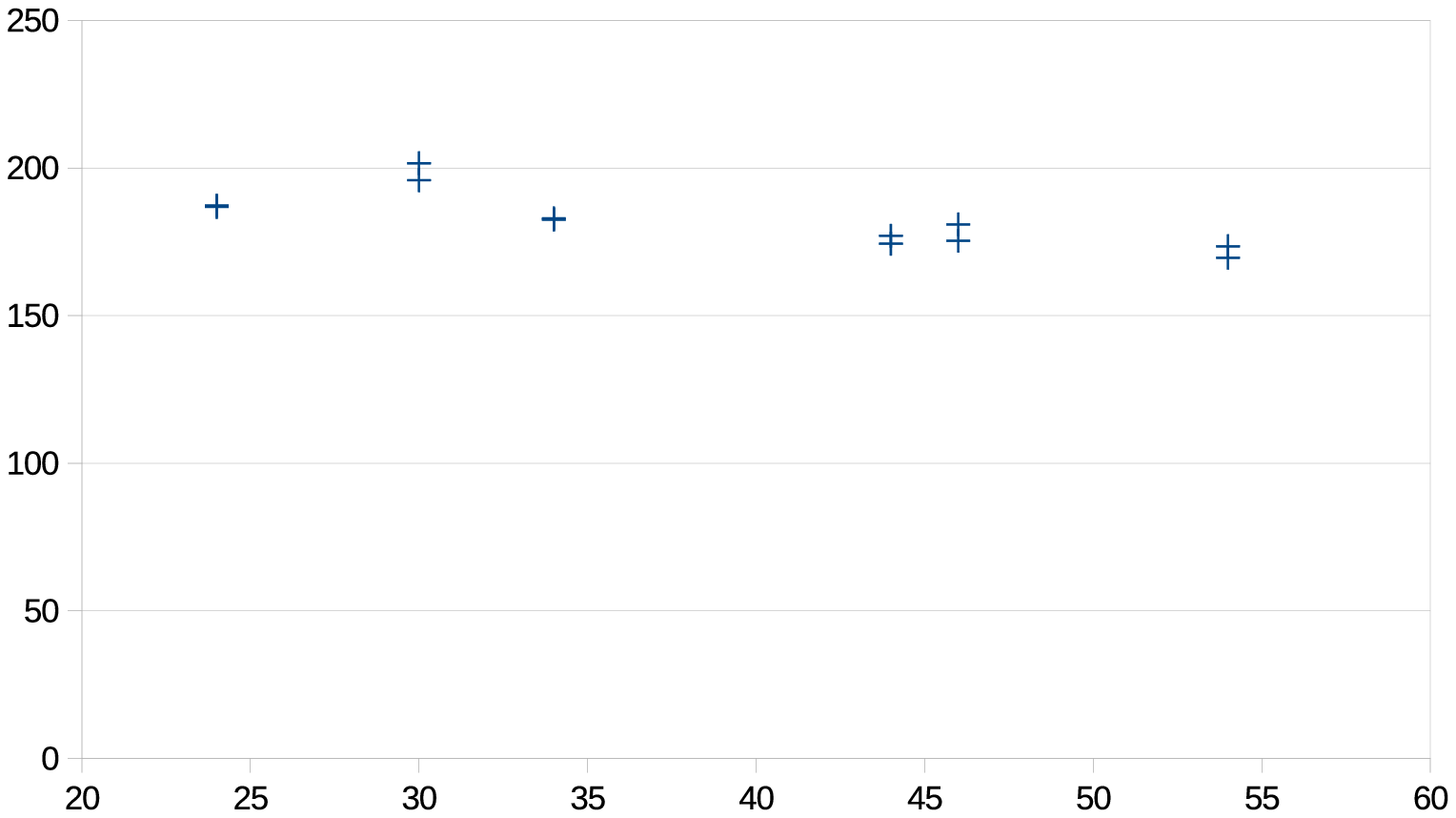}
}
\caption{\small{
(a) Squared difference leave-one-out value plotted as a function of number of network
parameters.
(b) Akaike Information Criterion, plotted against the number of free network parameters
(for; 2, 3, 4, 5, 2x2, and 2x3 hidden nodes).
}
}\label{fig:results}
\end{center}
\end{figure}

The cross-validation and AIC evaluation of selected neural architectures are
shown in Figure~\ref{fig:results}.
The entire dataset was run twice in each case from different
random number seeds in order to get a feel for the stability of these estimates.
The LOO estimate is absolute but less precise than the AIC. The AIC has the disadvantage
of requiring an estimate of the number of linearly independent parameters, which was
taken here to be the number of weights ($N_{\Theta}~=~W$) for simplicity.
Training a network with more than 55 weights was considered untenable given
only 118 samples.
The two methods broadly agree that there is no significant benefit in the
selection of any one specific architecture for this dataset.
Consequently, we choose the two extremes, 2 hidden nodes and two layer of three
hidden nodes, to illustrate the range of behaviours in tests.
These also happen to be the architectures which were marginally better
on one these two evaluation criteria.

\begin{figure}[ht]
\begin{center}
\subfigure[]
{
\includegraphics[scale = 0.5]{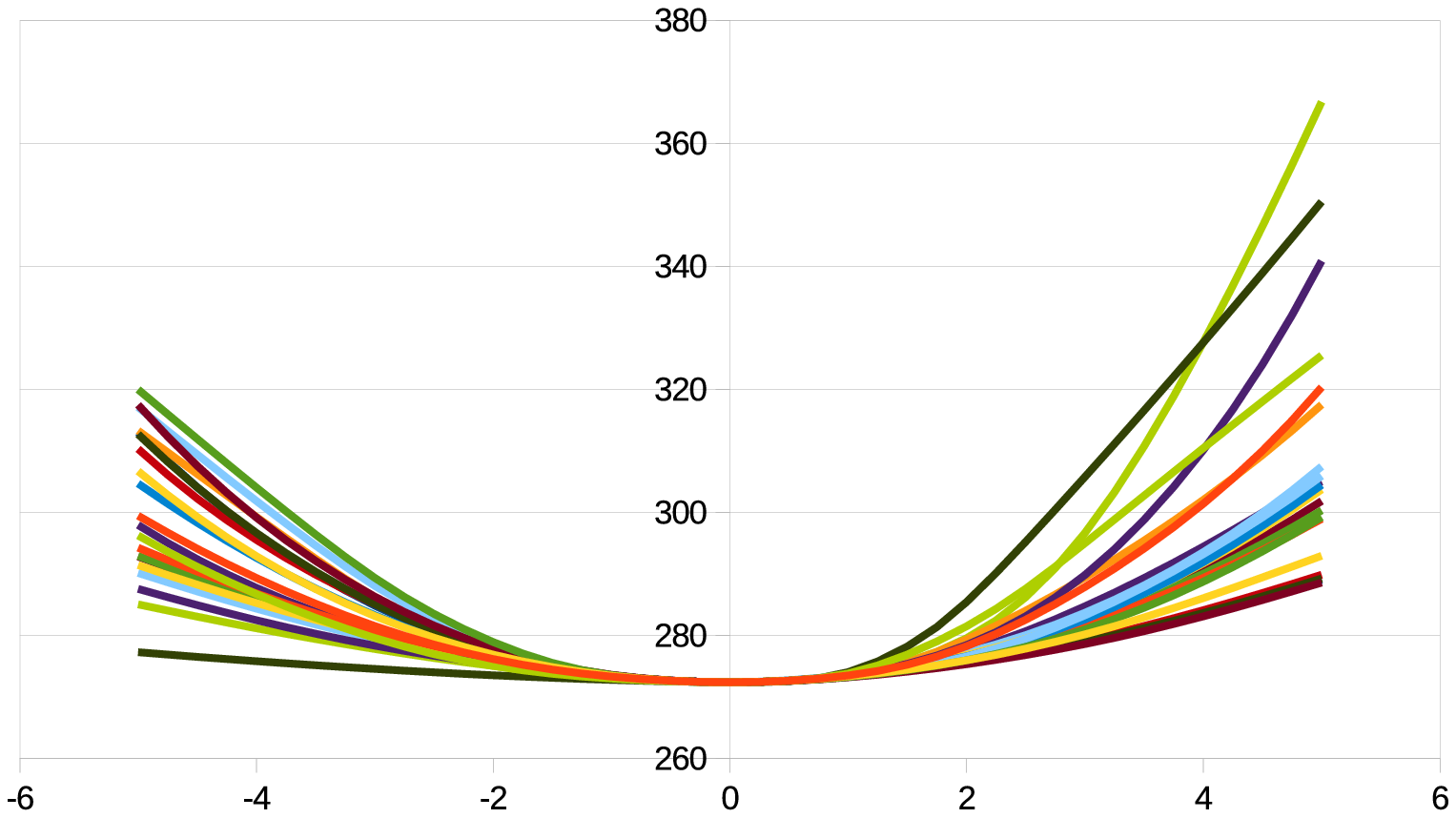}
}
\subfigure[]
{
\includegraphics[scale = 0.5]{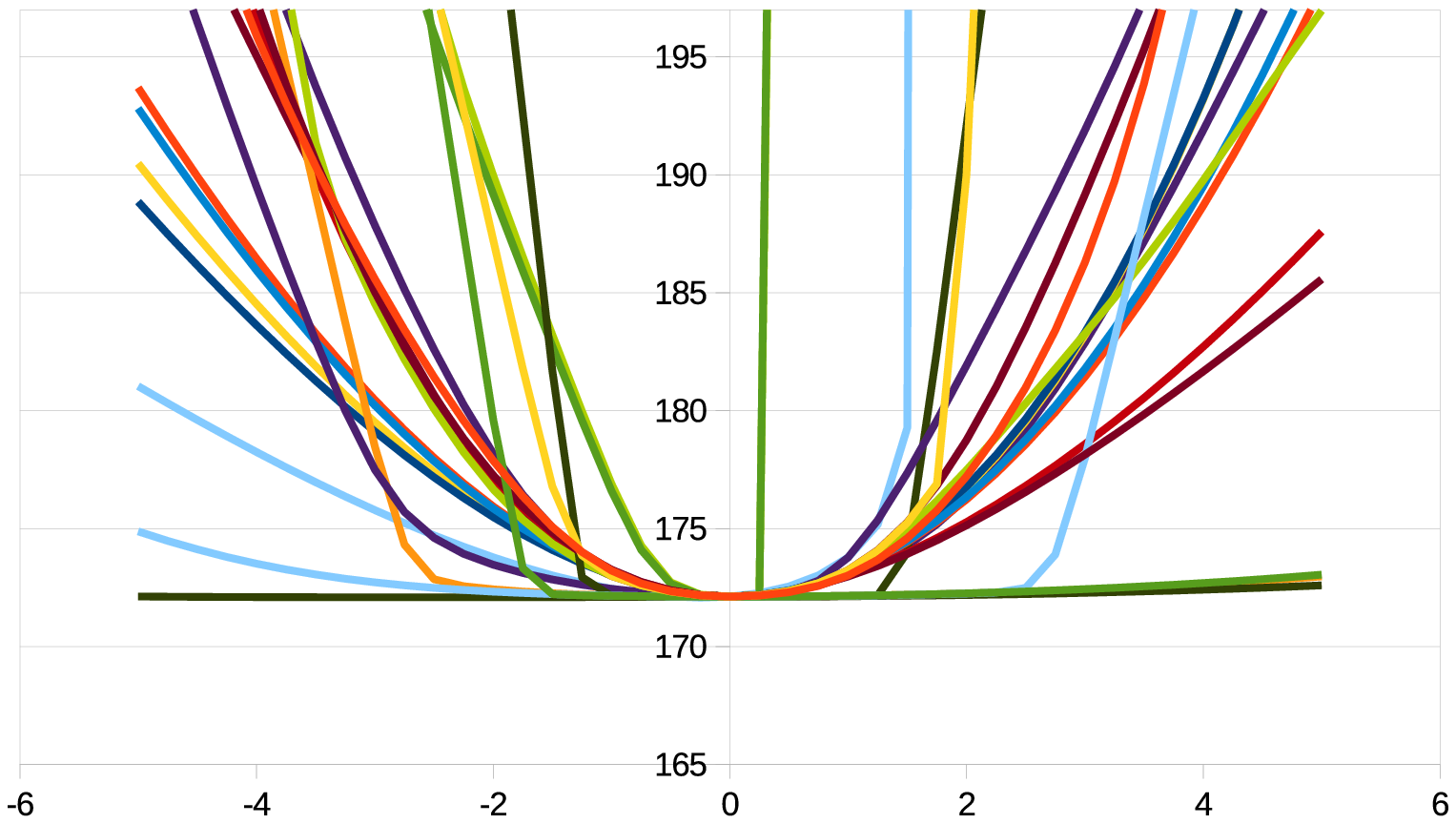}
}
\subfigure[]
{
\includegraphics[scale = 0.5]{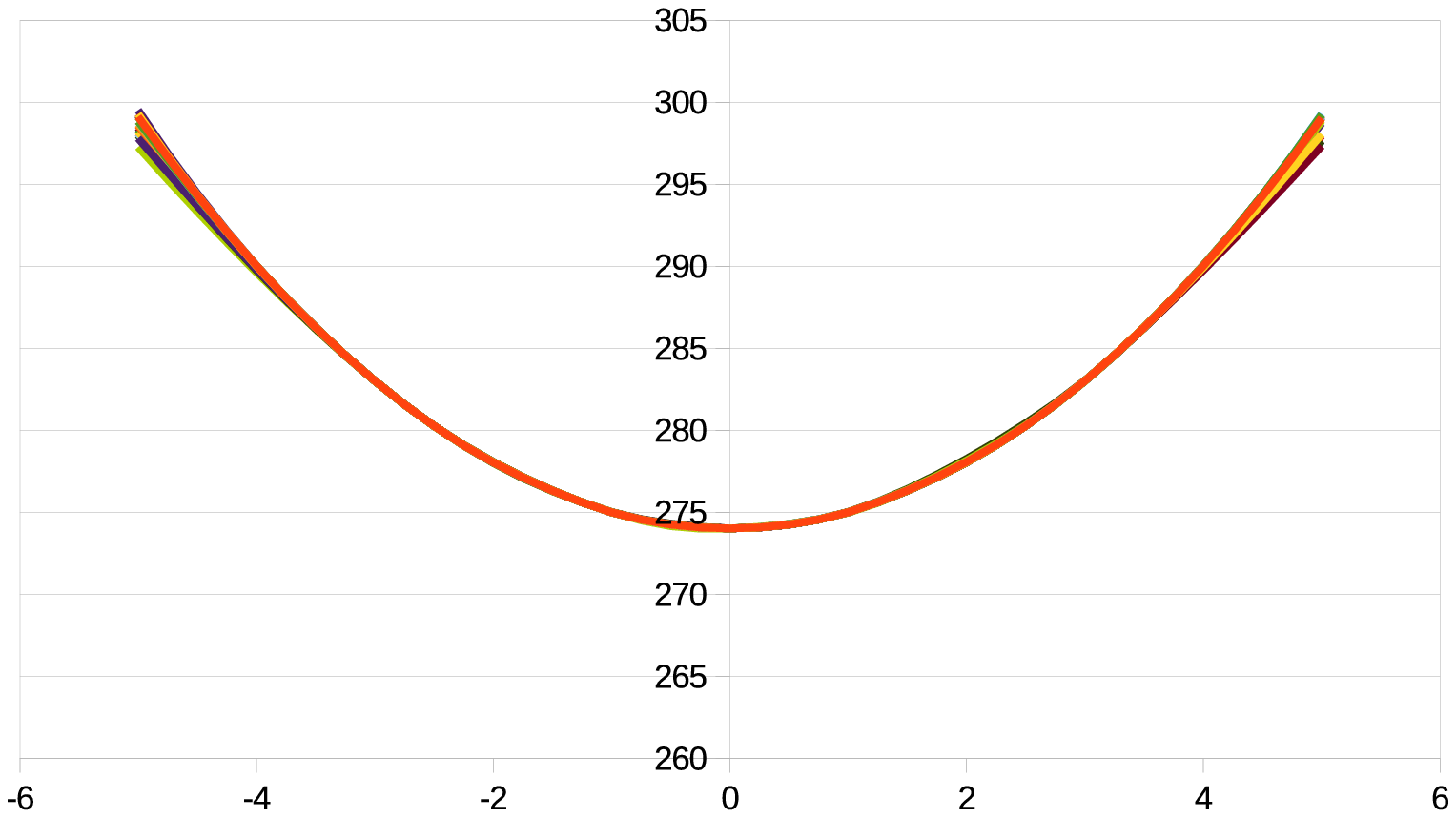}
}
\subfigure[]
{
\includegraphics[scale = 0.5]{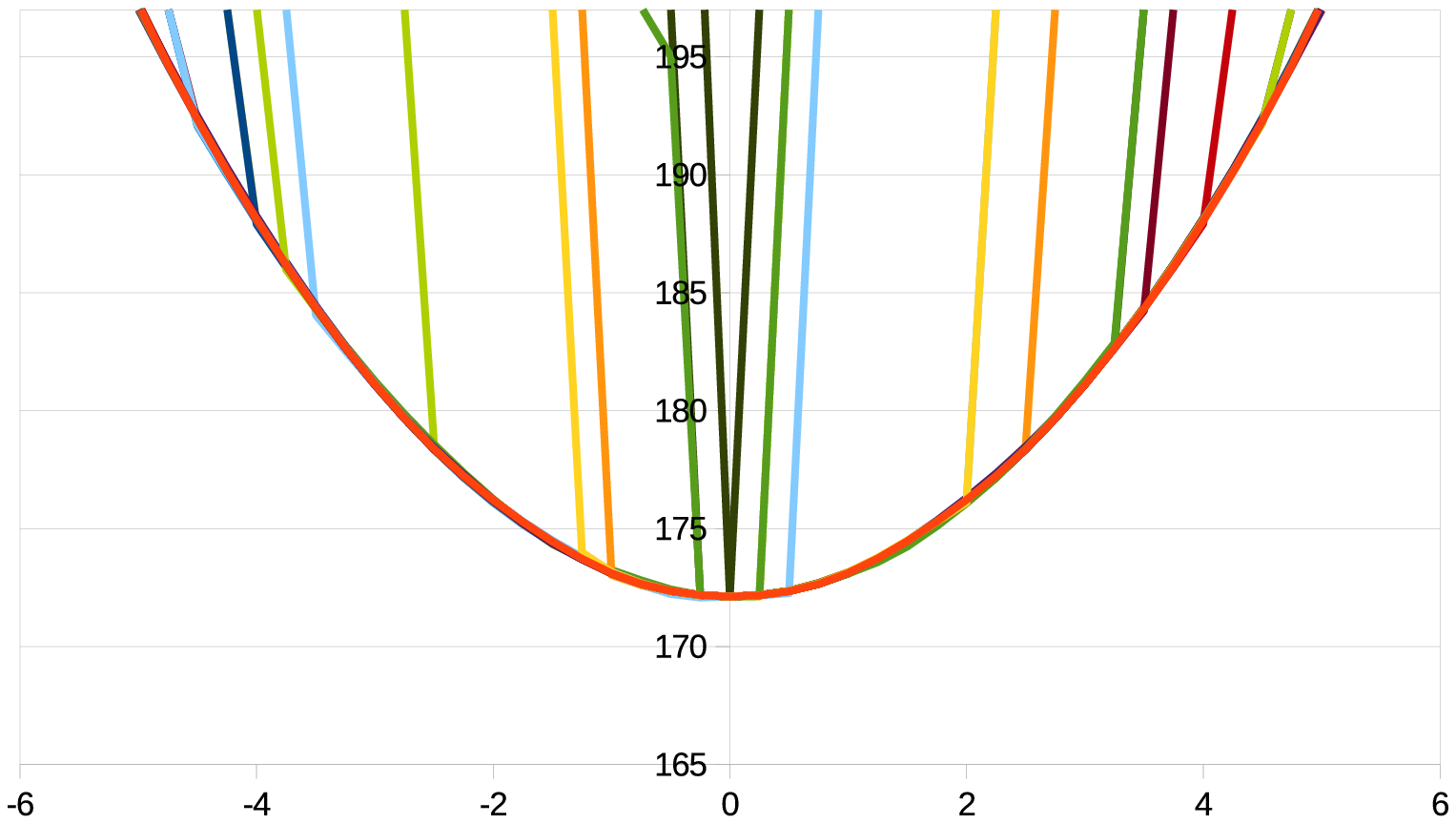}
}
\caption{\small{
(a) The log Likelihood function variation for 20 parameters form a network with 2 hidden nodes.
(b) The log Likelihood function variation for 20 parameters form a network with 2x3 hidden nodes.
(c) The log Likelihood function variation for 20 parameters form a network with 2 hidden nodes mapped to a quadratic.
(d) The log Likelihood function variation for 20 parameters form a network with 2x3 hidden nodes mapped to a quadratic.
The steep lines mark the maximum and minimum valid parameter range.
}
}\label{fig:weights}
\end{center}
\end{figure}

The re-mapping  of weight parameters is shown in Figure~\ref{fig:weights}. Prior to remapping the
log likelihood function is highly asymmetrical ((a) and (b)). The network with two hidden nodes
has all parameters constrained with a likelihood function which allows mapping on both sides
(Figure~\ref{fig:weights}(c)).
In contrast, the network with two layers of three hidden nodes has parameters which
do not map to the full range of $z$ (Figure~\ref{fig:weights}(b)). These are situations where changing the parameter
exhibits a plateau followed by a steep rise on one side of the optima. This is
likely due to the nature of the transfer functions,
which limit the potential for increase in a nodes output to a fixed value.
The addition of weights and nodes seems to increase the prevalence of this effect.
It is corrected for during uncertainty assessment by eliminating non-mapped ranges,
as described above.

\begin{figure}[ht]
\begin{center}
\subfigure[]
{
\includegraphics[scale = 0.5]{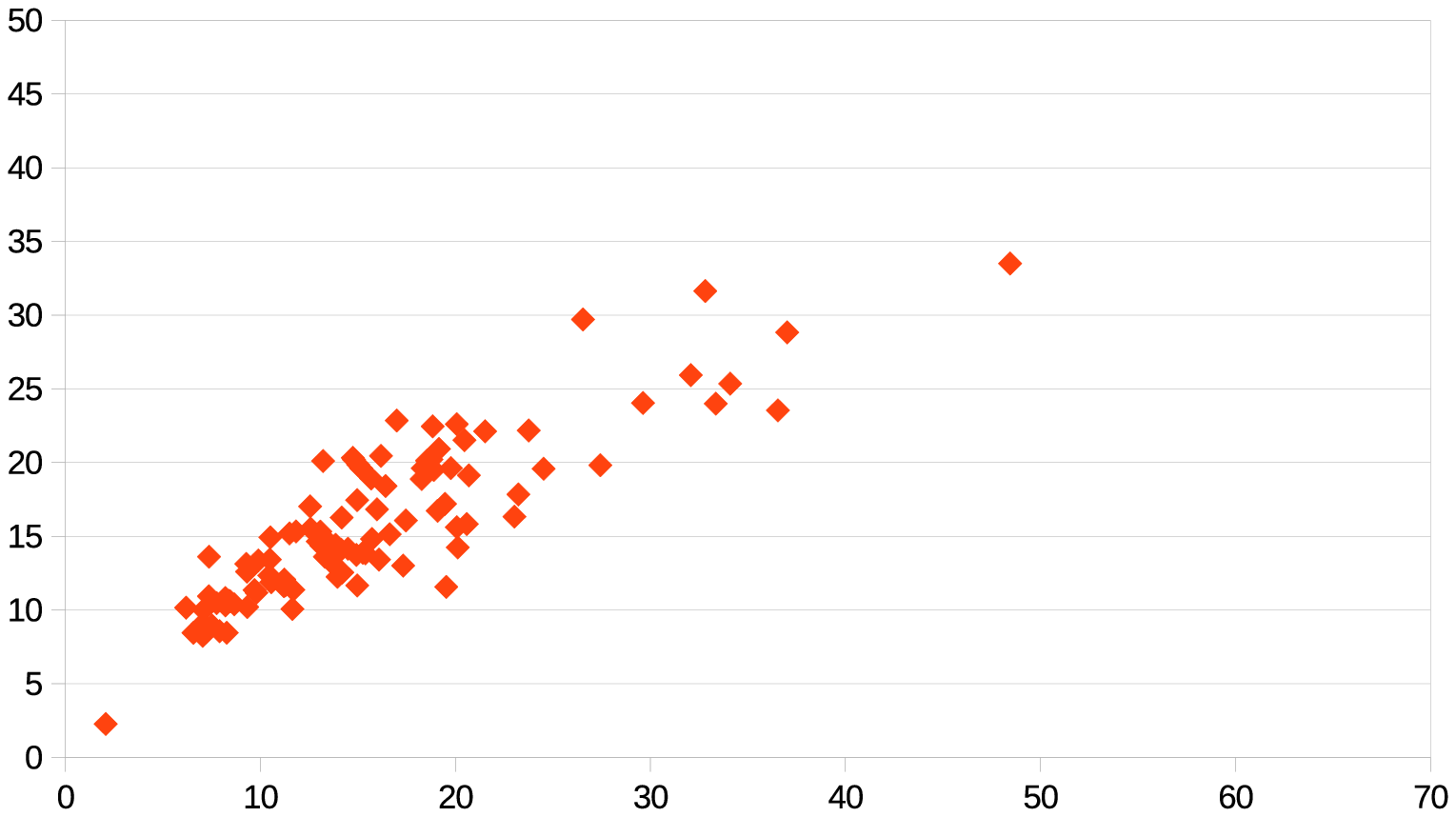}
}
\subfigure[]
{
\includegraphics[scale = 0.5]{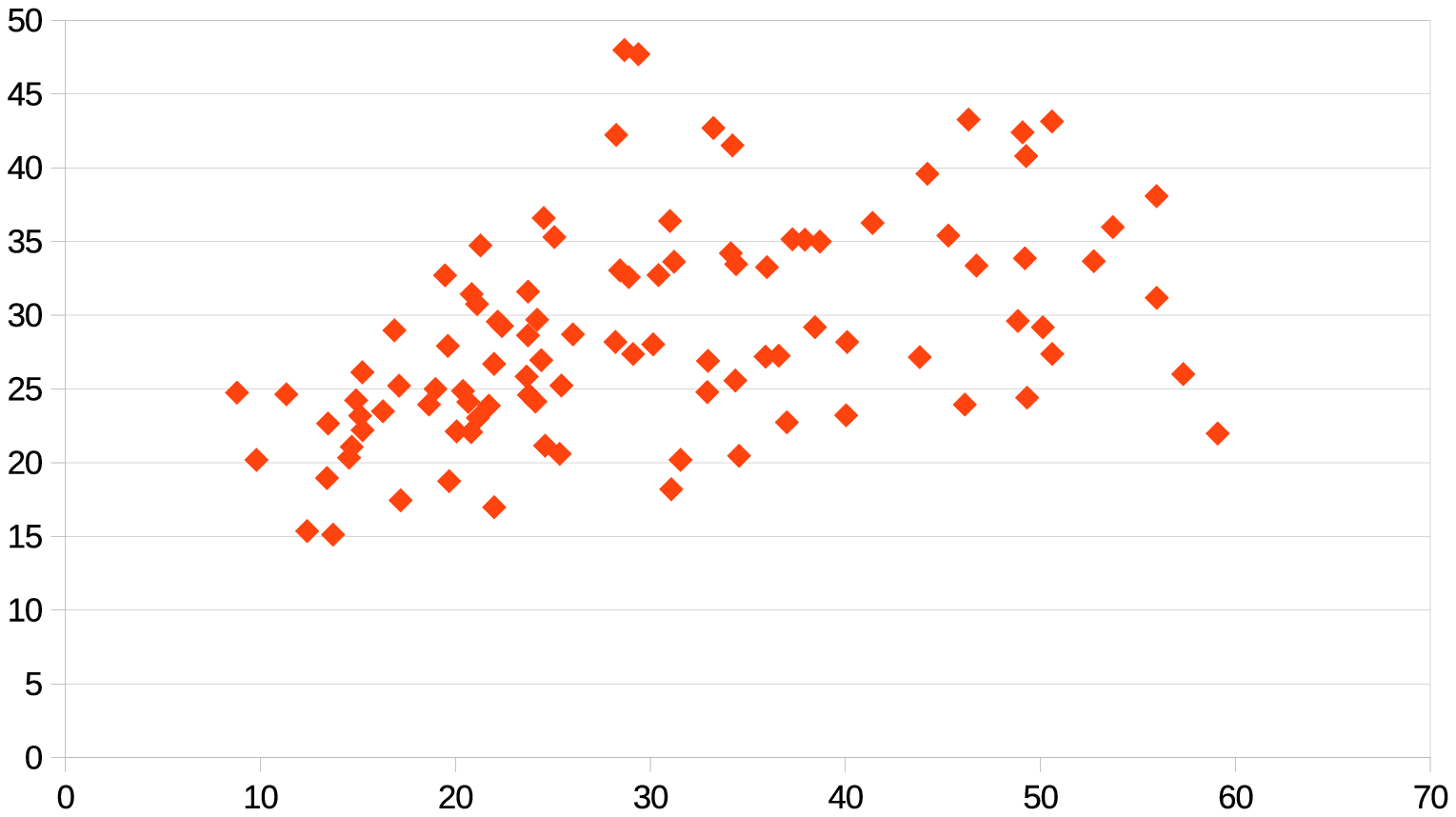}
}
\caption{\small{
(a) The distribution of Mahanaobis estimates compared against the original ANN cost function for
a network with 2 hidden nodes.
parameters.
(b) The distribution of Mahanaobis estimates compared against the original ANN cost function for
a network with  2 layers of 3 hidden nodes.
}
}\label{fig:Mahal}
\end{center}
\end{figure}

The distribution of Mahalanobis distance against original ANN cost function is shown in Figure~\ref{fig:Mahal}.
In theory these plots should be consistent with a line with unit slope. We can interpret spreading
around this line in terms as an equivalent random error on the estimate of each $z_w$.
A Monte-Carlo simulation for the network with two hidden nodes suggests that the amount
of variance around the expected line is consistent with a 0.1 random error on $z_w$,
i.e. it is statistically negligible given the expected accuracy of this parameter (i.e. 1).
The network with two layers of three hidden nodes exhibits both a greater variation
along the expect line (due to the change in the number of degrees of freedom) and greater
spread, now consistent with a random error of of 0.2. Although this is still
considered negligible, it provides an indication that as the complexity of the network
is increased, the degree of non-linearities increase and the available data are less
able to constrain parameter variation. As a consequence the degree of conformity of
the re-mapped parameter to a Gaussian Likelihood is reduced.

\begin{figure}[ht]
\begin{center}
\subfigure[]
{
\includegraphics[scale = 0.5]{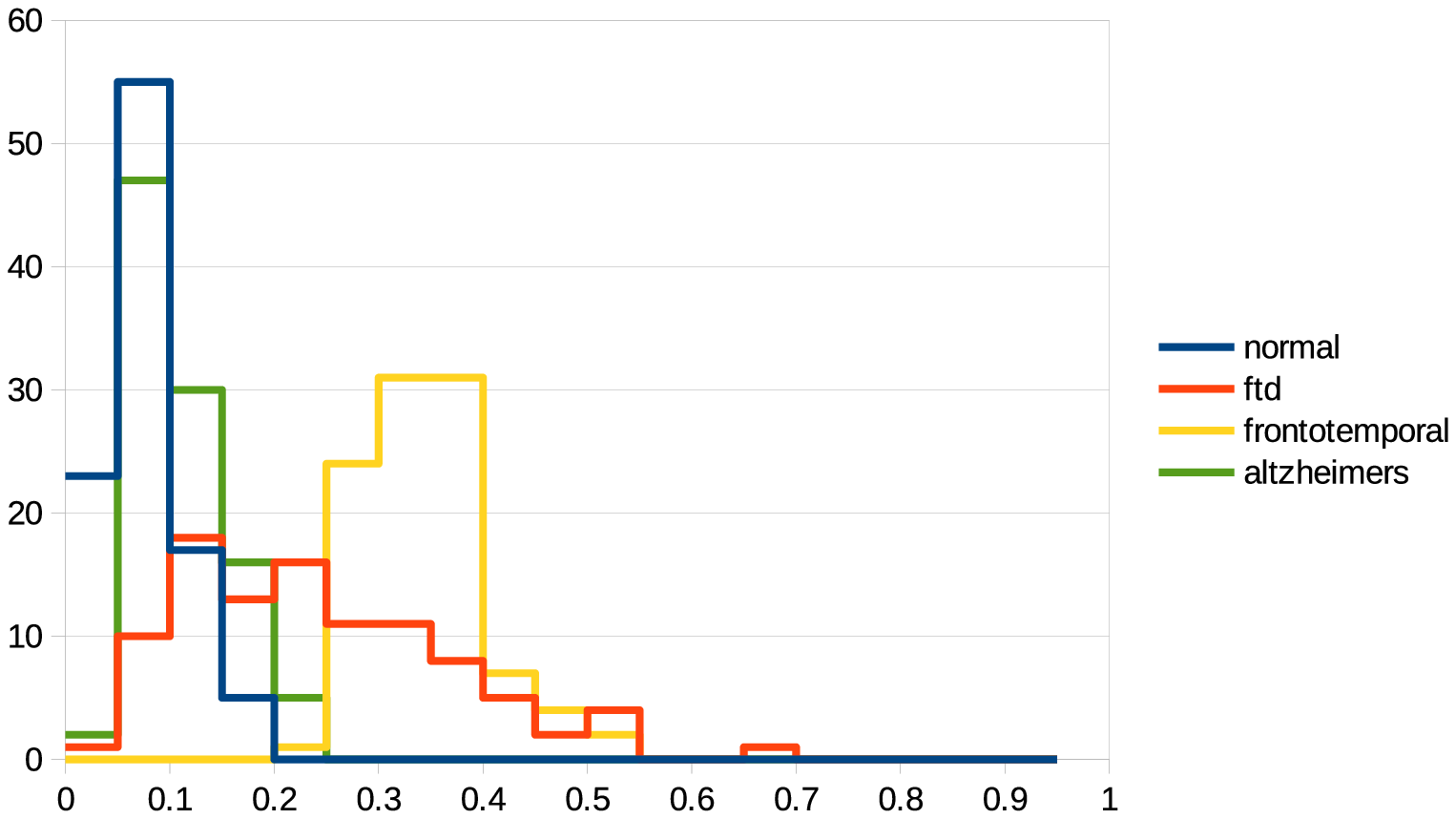}
}
\subfigure[]
{
\includegraphics[scale = 0.5]{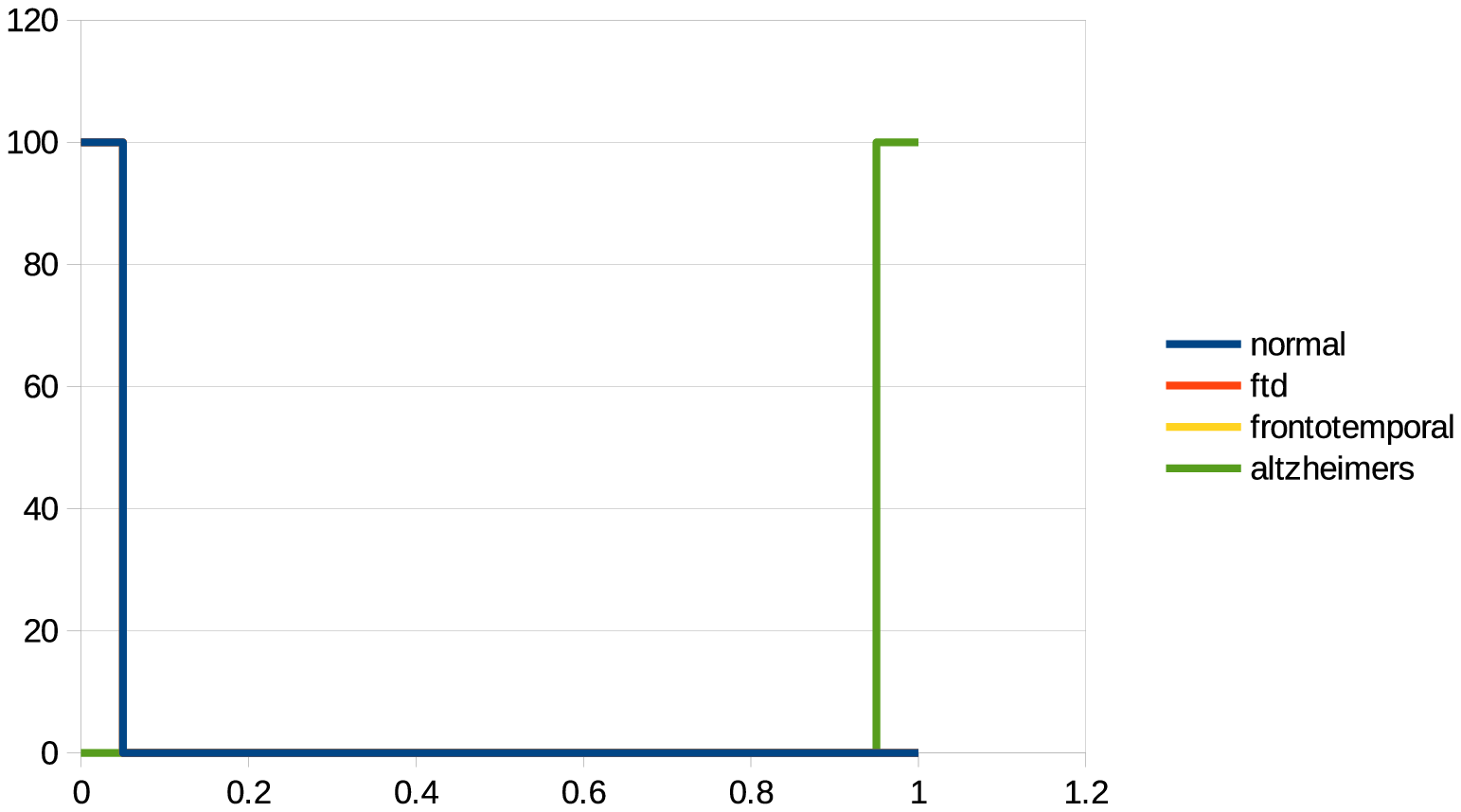}
}
\caption{\small{
(a) The frequency distribution of 100 classification outputs from a network with
2 hidden nodes computed by MCMC for
data point A in Figure~\ref{fig:reduced}(a).
(b) The frequency distribution of 100 classification outputs from a network with
2 hidden nodes computed by MCMC for
data point B in Figure~\ref{fig:reduced}(a).
}
}\label{fig:dist}
\end{center}
\end{figure}

The distribution over expected outputs for two patterns (A,B) identified
in figure 3 is shown in Figure~\ref{fig:dist}.
We can note that the uncertainty distribution for A reflects its position  among
the distributions seen in Figure~\ref{fig:reduced}(a). We will discuss the result for B below.

\section{Discussion}

\subsection{Numerical Considerations}

The noisy nature of the inverse covariance terms computed using equation
~(\ref{eqn:offdiag}) (typically a few percent), required a limit to be put on the absolute
values of off-diagonal terms to restrict them to be consistent
with the mathematical limit of unity. A truncation to $\pm 0.95$ was found to give
good stability without restricting the description of genuine parameter covariance.

The calculation of the Mahalanobis distance in Eqn.~(\ref{eqn:zprior}) required special
consideration. The often singular nature of the inverse covariance matrix
required the use of Singular Value Decomposition (SVD), and the calculation
of the Mahalanobis distance from its eigen vectors ($\bf e_k$) and
eigen values ($\lambda_k$), i.e.
\begin{equation}
{\bf z}^T C_{\bf z}^{-1} {\bf z} ~=~ \sum_k^W \frac{1}{\lambda_k^2} ({\bf e}_k.{\bf z})^2
\end{equation}
for use in Eqn.~(\ref{eqn:zL}). A relative minimum condition limit of
$ \lambda_k/ \lambda_{max} ~=~ 0.001 $ was also used, in order to
avoid large numerical instabilities.
Values below this value were set to this minimum limit value.

Finally, in order to restrict the samples to a more realistic approximation of
the inverse parameter covariance, a maximum value was put on the true ANN
cost function (equivalent to a chi-square of $4N_{\Theta}$) as
each sample was generated, in order to eliminate implausible
instances of weights.

\subsection{Data Representation and the Use of Prior Knowledge.}

Although the original published work ~\cite{Thacker} might be useful as a bench-mark,
it was not the intention here to perform a shoot-out between
ANNs and kNN. Given the small sample size we are highly restricted in our
choice of ANN architectures and it could not be done with any statistical power.
Rather, the purpose of the current work is to illustrate how the epistemic (systematic) uncertainties in
ANN outputs, due to the original uncertainties in training data samples, can be estimated
for safety critical applications.

However, this work does identify some
important issues regarding the use of prior knowledge and low parameter pattern
recognition solutions to mitigate against problems with ``black box'' approaches ~\cite{Ramesh}.
It must be stated here that for scientific and clinical studies requiring
novel imaging there will always be difficulty in generating unlimited
data quantities, due to the need for specialised ethics approval and funding.

Regarding the use of prior knowledge, volume normalisation was performed on the
basis that the proportional structure of brains is expected to be independent of overall
size, and specific diseases tend to affect specific anatomical regions. It is
also known that normal ageing causes a gradual monotonic decrease in brain
tissue volume after the age of about 40. Age corrections were performed with the
intention of making the distributions of normal and disease groups less ambiguous.

Dimensional reduction was achieved by summing the individual (12) measured volumes into
variables which might be expected to correlate due to proximity: front (4), middle (4),
back (4), left (6), right (6), top (6) and bottom (6). From them some diagnostically relevant,
mathematically orthogonal and homoscedastic combinations were constructed $X_i$ (see Appendix~\ref{app2}). These combinations were
identified by a subjective observation of graphed variables (e.g. Figure~\ref{fig:reduced}), in order
to identify those which illustrated some obvious degree of class separability.
Importantly, we did not set disambiguation as a quantitative target, as we wanted
to reduce the chances of over-fitting, although the general effect of ambiguity
reduction was confirmed graphically.

We have not enforced a Bayesian re-normalisation of the output probability
estimates, as despite attempts to identify subjects with unique diagnoses,
we decided that in-use the classifications would not be
mutually exclusive, making Bayes Theorem in-appropriate.
However, we would otherwise expect that exploitation of this constraint would
improve the effective DOF in training data.

In the original work ~\cite{Thacker}, the final five homoscedastic variables ($X_1,.. X_5$) were re-scaled, using
the available repeated measurement information, to approximate measurement
precision. This last step was considered particularly important when using
the variables in a kNN, as it imbued the Euclidean distance with a nominal
statistical scaling. We omit this scaling in this work, as the extra
parameters in an ANN can be adjusted to achieve this. However, the
conventional ANN training does
not exploit our knowledge regarding repeat precision and
generates more parameter complexity than the corresponding kNN.

Whilst the corrections for volume and age, along with dimensional reduction to
independent homoscedastic variables, and even Bayesian re-normalisation,
could have been attempted by adding extra layers to the ANN (e.g. deep learning),
the extra degrees of freedom and associated non-linearities could not have
been exactly replicated with standard transfer functions and network architectures.
Also the specific details of how to leverage the subsets
of data (i.e. normals and repeated measurements for these processes) to
achieve stable corrections, could not have been determined in a bottom-up
manner without having access to exponentially more quantities of data
(a 14 dimensional pattern space density rather than 5).

In summary, ignoring the prior knowledge, and attempting to replicate
the pre-processing using trained neural network calculations (extra layers), would have
needlessly put us close to the regime of deep learning and big data.
As we have already stated above, there are reasons why
such data is not going to be available in these kind of studies.
Even if the data is available there are still benefits to good choice
of input data representation.
A low dimensional space can be mapped more accurately than
a higher one containing the same information ~\cite{Lacey} and we can guarantee that
the calculations performed (such as scale invariance, variable normalisation
and age correction) are appropriate and unconditionally stable across the data space.
The final representation variables also support clinical interpretation
by being related to simple structural biology. Such simple transparency was
considered an important requirement for future clinical integration
~\cite{Leslie}.

\subsection{Implications for Assessment of ANN Uncertainty}

ANNs pose a particular challenge for the assessment of parameter uncertainty.
The conventional approach, based upon second derivatives ~\cite{Murata},
is unlikely to generate useful estimates of the inverse covariance matrix, due to the
strong non-linearities used in ANN calculations. Whilst re-mapping the
parameters to achieve a Gaussian likelihood function improves the  inverse
covariance approximation to the original cost function, and this alone may well be
a good enough reason to take the approach, this is not the only reason
for doing this. As we have explained in the introduction, making the Likelihood function
Gaussian is the Frequentist equivalent to the Bayesian use of
uninformative ``priors'', the multiplicative terms needed to convert
the likelihood into a quantitatively valid density estimate.
The plots in Figure~\ref{fig:Mahal} are therefore as much a test of the probability theory
as they are a test of the accuracy of the inverse parameter covariance.
This also suggests that we cannot improve the conventional methods using higher order statistics
to better the model the likelihood function, as the estimates of parameter
density arising from the theory are simply wrong unless the likelihood is
Gaussian, i.e. we cannot treat a likelihood function as though it is already a density over
parameters.

At first glance the estimation of uncertainty shown for pattern B in Figure~\ref{fig:dist}(b)
might seem completely reasonable when we take into account its location in Figure~\ref{fig:reduced}(a).
However, this location corresponds to a very low sample density, much like the
data for the binomial example shown in Figure~\ref{fig:Ex2}(a), but does not have the same degree of variation.
It is more tightly grouped around $P(C|{\bf X})=1$ because the ANN is a parametric fit.

\begin{figure}[!b]
\begin{center}
\subfigure[]
{
\includegraphics[scale = 0.5]{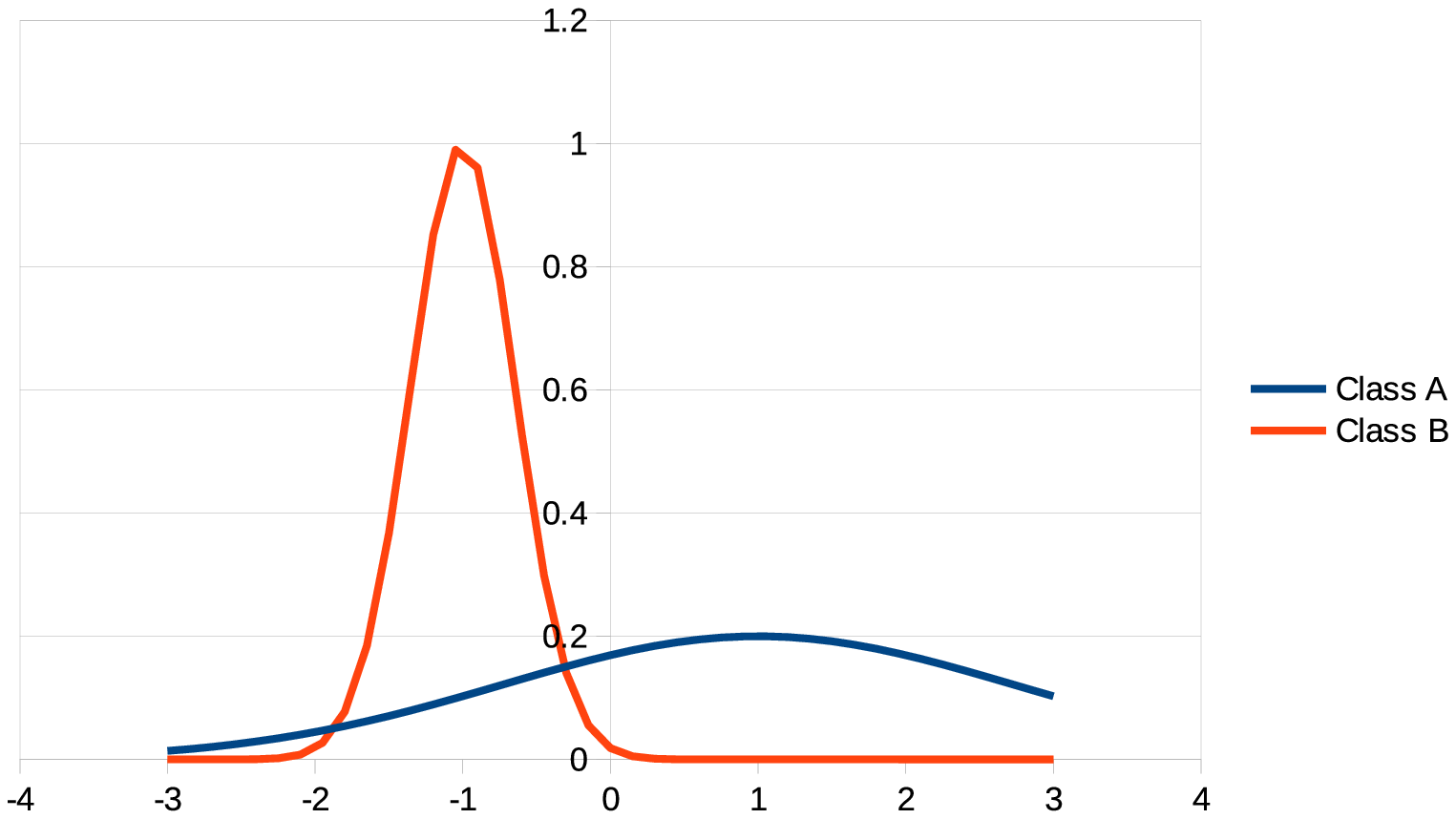}
}
\subfigure[]
{
\includegraphics[scale = 0.5]{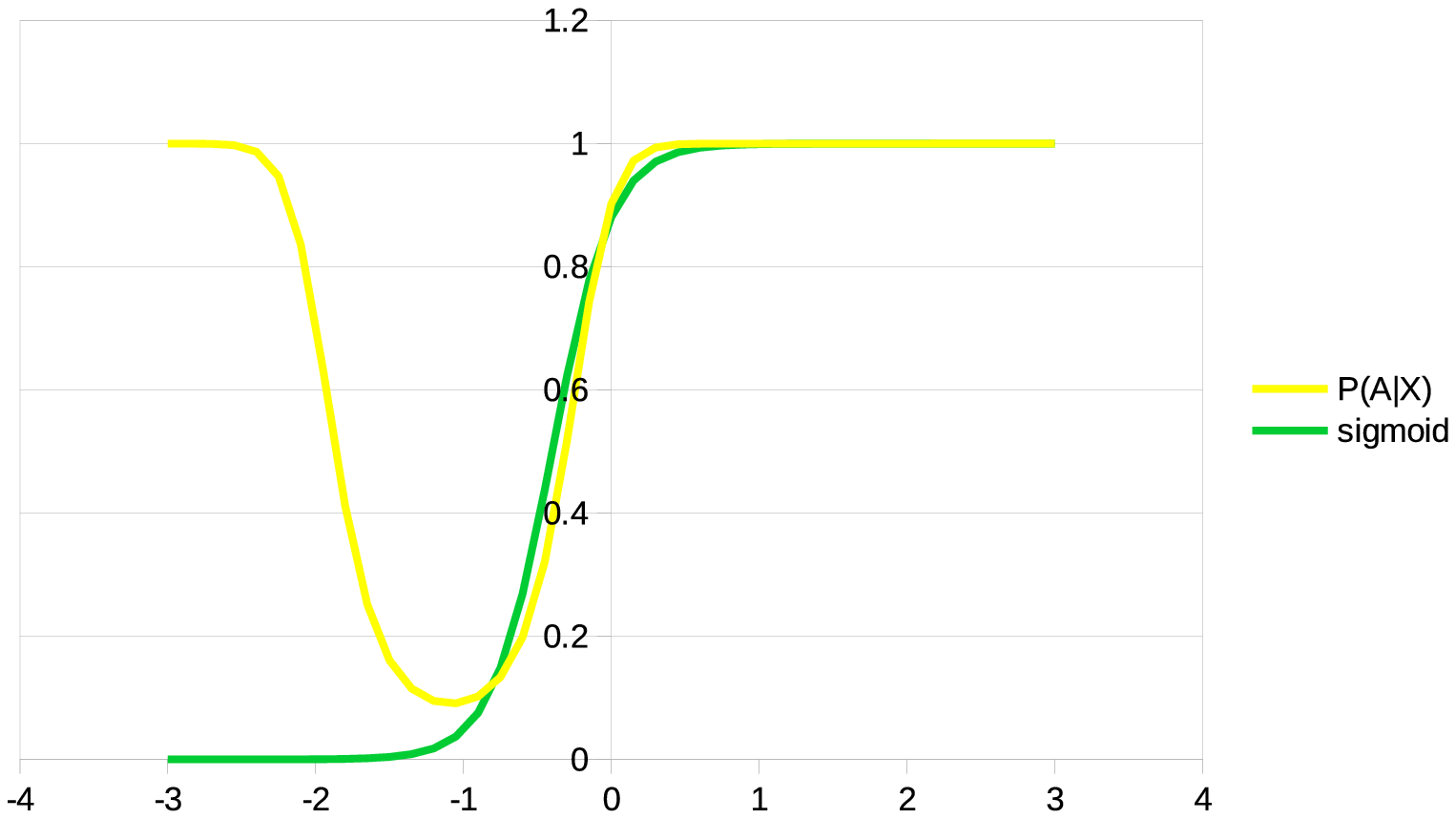}
}
\caption{\small{
The conditional probability of classification (yellow) is computed for two
classes (A blue and B red) with Gaussian distributions of unequal width.
An ANN trained with small numbers of samples from these distributions
is likely to model the decision boundary using a sigmoid (green).
Future test samples for $X < -2$ will then be classified
in the same incorrect way, regardless of any expected uncertainty
in the location or scale of the sigmoid.
The estimated epistemic error for such data
(a few percent of all samples) would approach zero,
even though the decision is always categorically wrong.
}
}\label{fig:gauss}
\end{center}
\end{figure}

Sigmoid functions have no choice but to be stable at either extremes
of the input variables, where the outputs are $(0,1)$.
Unfortunately, this behaviour  may not necessarily be justifiable.
For example, if we consider the very common assumption that the underlying class distributions
are Gaussian, then for two one dimensional densities
with non-equal width, the conditional probability
would converge on outputs for the broader distribution of $(1,1)$ (Figure~\ref{fig:gauss}).
These issues are not unique to ANNs, they also occur in familiar statistical methods,
for example both linear and logit regression are capable of generating nonsense if misused. We may be satisfied to
make these assumptions where we know something about the input data properties.
Biological factors regarding changes in brain structure due to disease
(loss of tissue in specific anatomical regions) would lead us
to expect the diagnostic classification would  be as modelled here. However,
if this were a ``black box'' application, with the same distribution of training data,
we would have no right to just expect such assumptions to be true{\footnote{This example appears to us to contradict the assertion that uncertainties for mis-specified models
can be accommodated when using NIC ~\cite{Anders}.}}.

At this point it is instructive to consider a simpler pattern recognition system, called Linear Poisson Modelling (LPM) ~\cite{Tar}, to emphasise the importance of model selection and quality control.
LPMs approximate probability densities non-parametrically (i.e. with a minimum of
functional restrictions) for systems which are {\bf known} to be
modellable as linear combinations of probability mass functions. The associated
``white box'' analysis ~\cite{Ramesh} supports estimation of both epistemic and aleotoric
parameter uncertainties.
To be applied correctly, the theory makes it clear that training data must take
the form of histogram bins with independent Poisson samples. The uncertainties on model coefficients are computed using error propagation, assuming that sufficiently large Poisson sample perturbations can be approximated with a Gaussian covariance matrix. These assumptions can be, and often are, violated in real data.
Consequently, whilst histograms appear ubiquitously within science, it has proven challenging to apply LPMs `out-of-the-box'. A mass spectrum, for example, is a type of histogram for ion counts.
However, what is recorded are noisy step-changes in voltages subjected to non-Poisson instrumentation noise.
LPMs have been successfully applied to such data ~\cite{Deepaisarn}, but only with significant pre-processing and calibration ~\cite{Seepujak}.
Tools such as Bland-Altman analysis are needed to confirm noise distributions; pull-distributions check that predicted uncertainties match those observed; and simulation and ground-truth datasets are used to corroborate all model assumptions.

Without a linear data generator and statistical conformity, predicted uncertainties on LPM derived measurements can be orders of magnitude away from reality.
The current trend within ANN research has been to present raw data and tune parameters until an ``acceptable'' empirical result has been achieved on a test dataset.
We should not expect that a ``black box'' ANN would reliably
identify and solve complex problems within data,
especially when a comparatively simple method requires such careful application.

Fundamentally our statistical analysis of uncertainty is based upon the
assumption that the implied non-linear ANN model is valid, when
in fact it is only an interpolating approximation of the training data. If we cannot trust
the non-linear model then the computed uncertainties away from training data
represent a best case, i.e.  a  statistical variation around a biased output.
In safety critical situations we might be better using less stable parametric
(or non-parametric)
models so that uncertainty estimates reflect the limiting information content of training data.
The requirement to understand uncertainty in saftey critical tasks therefore has implications
for how we might choose ANN architectures.

\subsection*{General Conclusions}

Regarding the general theory of uncertainty, standard results
for Jeffreys priors give near identical results to the Frequentist approach of
mapping the likelihood function onto a Gaussian. The main difference however,
is that whilst general solutions to the construction of Jeffreys priors are
highly complicated and impractical for complex non-linear systems, we have shown in
this work that an approximation to a general solution for the Frequentist
approach is numerically feasible. It also embodies a pragmatic technique
for the better  approximation of uncertainties using covariance matrices.

Now that we have a working system for the estimation of epistemic
ANN output uncertainty we can deduce several fundamental issues
regarding the feasibility of obtaining a successful uncertainty assessment:

\begin{itemize}
\item In order to apply the parameter-remapping technique to the estimate
of parameter uncertainty ($p(\Theta| {\bf X})$), the likelihood function
needs to be fully optimised. Although it is possible to deal with
a cost function which has a plateau on one side of the optima, a partially
optimised function can provide no information regarding the
constraint (estimation error) on the associated parameter. Such
issues have been reported previously ~\cite{Anders}.
This has implications for training methods which employ early stopping
i.e. it will be logically impossible to make any meaningful prediction regarding uncertainty.

\item In order to compute numerically stable estimates of the Mahalanobis
distance, it is necessary to apply SVD for the extraction of
eigen-vectors from the re-mapped parameter covariance.
It may be impossible to estimate the Mahalanobis distance to sufficient
precision to work with even 100's of network parameters unless
some other way can be found either to deal with ill-conditioned matrices
or to remove parameter correlation from the ANN design.

\item The mapping process is also best if the cost function is smooth
and differentially continuous, as discontinuities could pose numerical problems.
This has implications for common processing acceleration systems,
involving the use of RELU transfer functions.

\item The mapping approach also has more difficulty as the degree of non-linearity
is increased, such as the process of adding additional non-linear
hidden layers to the ANN. Also, as the architecture complexity is increased
the increased number of degrees of freedom reduces
the average statistical information available to constrain parameters (as
described in ~\cite{Murata}).
This in turn accentuates any non-linearities, by allowing the expected parameter
variations to span a larger range of values. As a consequence, large dimensional
input spaces and small datasets, which lead to over-fitting, will reduce
the accuracy of any Gaussian approximation.
This has implications for the assessment of uncertainty when using deep-learning.

\item Any statistical assessment of  uncertainty is based upon the implicit
non-linear ANN model being valid. As this can not generally be expected to be true
in areas away from training data
the degree of variation estimated can strictly only be a lower bound.
\end{itemize}

The above issues have obvious implications for the application of deep-learning
to safety critical tasks. However, they can be mitigated to some extent
by reducing the need for large
networks by choosing appropriate input data representations, rather than simply
using raw data, much as was common practice ~\cite{Lacey,Thacker, Bromiley} before the advent of Bigdata.








\bibliographystyle{abbrv}
\bibliography{Likelihoods}

\appendix
\section{Iteration of Jeffreys Priors}\label{app1}

In order to obtain the matching equation for the prior, we assumed that we could manipulate the prior in order to obtain a better fit near the optimum between the posterior and a unit Gaussian. The result shows that the Jeffreys prior gives the \emph{first-order} approximate answer to mapping the Likelihood to a Gaussian. The Jeffreys prior can hence be used to define a mapping of (log)-likelihood functions for some range of parameter values about an optimum thus:
\[
\theta \mapsto \omega(\theta) \:\: \& \:\: l(\theta) \mapsto \widetilde{l}(\omega) \:\: \mbox{where:} \: \: \widetilde{l}(\omega(\theta)) = l(\theta), \:\: \left(\frac{d\omega}{d\theta}\right)^{2} \doteq I(\theta) \equiv -\frac{d^{2}l(\theta)}{d\theta^{2}},
\]
where we now consider the total (log) Likelihood and the total Fisher Information function. This Jeffreys-prior-based mapping procedure gives us an order-preserving and optimum-preserving mapping between Likelihood functions. The derivation above showed us that \emph{one} application of this Jeffreys mapping gave us a better approximation to a Gaussian Likelihood. It is straightforward to see that a unit Gaussian is a \emph{fixed point} of this Jeffreys mapping, since:
\[
l(\theta) = -\frac{1}{2}(\theta-\MLE)^{2} \:\: \Rightarrow \:\: I(\theta) \doteq -\frac{d^{2}l(\theta)}{d\theta^{2}} = 1 = \left(\frac{d\omega}{d\theta}\right)^{2}\:\: \therefore \:\:\omega(\theta) = \theta.
\]
If we take a Gaussian with a different scale:
\begin{align}
l(\theta) &= -\frac{1}{2\sigma^{2}}(\theta-\MLE)^{2} - \frac{1}{2}\ln (2\pi\sigma^{2}), \nonumber \\
\Rightarrow \:\: I(\theta) &= \frac{1}{\sigma^{2}} \:\: \Rightarrow \:\: \frac{d\omega}{d\theta} = \frac{1}{\sigma} \:\: \Rightarrow \:\: \omega(\theta) = \frac{\theta}{\sigma}, \nonumber \\
\therefore \:\: \widetilde{l}(\omega) &= -\frac{1}{2}(\omega-\hat{\omega})^{2} - \frac{1}{2}\ln (2\pi\sigma^{2}),
\end{align}
so we see that it maps to a unit Gaussian after one iteration.

Let us consider a perturbation to a Gaussian. In terms of the log-likelihood, we consider a cubic perturbation to the quadratic Gaussian log-likelihood. After some algebra, we find that:
\begin{align}
\mbox{If:} \:\: l(\theta) &= -\frac{1}{2}(\theta-\MLE)^{2} + \epsilon (\theta-\MLE)^{3}, \:\: |\epsilon| \ll 1, \nonumber \\
I(\theta) &= 1 - 6\epsilon(\theta-\MLE) \:\: \Rightarrow \:\: I^{\frac{1}{2}}(\theta) = \frac{d\omega}{d\theta} = 1 - 3\epsilon(\theta-\MLE) + O(\epsilon^{2}), \nonumber \\
\Rightarrow \:\: \omega(\theta) &= \theta - \frac{3\epsilon}{2}(\theta-\MLE)^{2} + O(\epsilon^{2}) \:\: \Rightarrow \:\: \theta = \omega + \frac{3\epsilon}{2}(\omega-\hat{\omega})^{2} + O(\epsilon^{2}). \nonumber \\
\therefore \:\: \tilde{l}(\omega) &=  -\frac{1}{2}(\omega-\hat{\omega})^{2} -  \frac{\epsilon}{2} (\omega-\hat{\omega})^{3} + O(\epsilon^{2}).
\end{align}
This means that in terms of the log-likelihood function, the magnitude of the coefficient of the perturbation term has been halved $\epsilon \mapsto -\epsilon/2$. We hence see that the Gaussian fixed point for the Jeffreys mapping has some non-empty basin of attraction.

\section{Dimensionality Reduction for Volumetric MR Data}\label{app2}

It was assumed that the
normal development of atrophy in any normalised and pooled variables $V$ would be
of the form
\[
V ~=~ (age ~-~ C)/K
\]
Thus we can construct a new variable representing the proportion
of atrophy relative to that expected at a particular age.
\[
V' ~=~ V K/(age ~-~ C)
\]
Variables which combined Anscombe transforms ~\cite{Anscombe1948} and partial orthonormality were
selected as follows;

\begin{itemize}
\item The age corrected relative degree of atrophy between the
middle and front of the CSF space.
\[
 X_1 ~=~ ( \sqrt{M'} - \sqrt{F'}) / \sqrt{2}
\]
\item The age corrected relative degree of atrophy between the
middle back of the CSF space.
\[
 X_2 ~=~ ( \sqrt{M'} - \sqrt{B'}) / \sqrt{2}
\]
\item The age corrected relative degree of total atrophy
\[
 X_3 ~=~ ( \sqrt{F'} + \sqrt{M'} + \sqrt{B'} ) / \sqrt{3}
\]
\item The age corrected relative degree of atrophy between the
left and right
sides of the CSF space.
\[
 X_4 ~=~ ( \sqrt{P'} - \sqrt{S'}) / \sqrt{2}
\]
\item The age corrected relative degree of atrophy between the
top
and bottom of the CSF space.
\[
 X_5 ~=~ ( \sqrt{U'} - \sqrt{L'}) / \sqrt{2}
\]
\end{itemize}

\end{document}